\documentclass[11pt]{article}

\usepackage[utf8]{inputenc}
\pdfoutput=1
\usepackage[margin=1.1in,letterpaper]{geometry}
\usepackage[round]{natbib}
\usepackage{tikz}

\makeatletter
\newlength\aftertitskip     \newlength\beforetitskip
\newlength\interauthorskip  \newlength\aftermaketitskip

\def\maketitle{\par
 \begingroup
   \def\thefootnote{\fnsymbol{footnote}}
   \def\@makefnmark{\hbox to 4pt{$^{\@thefnmark}$\hss}}
   \@maketitle \@thanks
 \endgroup
\setcounter{footnote}{0}
 \let\maketitle\relax \let\@maketitle\relax
 \gdef\@thanks{}\gdef\@author{}\gdef\@title{}\let\thanks\relax}

\def\@startauthor{\noindent \normalsize\bf}
\def\@endauthor{}
\def\@starteditor{\noindent \small {\bf Editor:~}}
\def\@endeditor{\normalsize}
\def\@maketitle{\vbox{\hsize\textwidth
 \linewidth\hsize \vskip \beforetitskip
 {\begin{center} \LARGE\@title \par \end{center}} \vskip \aftertitskip
 {\def\and{\unskip\enspace{\rm and}\enspace}%
  \def\addr{\small\it}%
  \def\email{\hfill\small\tt}%
  \def\name{\normalsize\bf}%
  \def\AND{\@endauthor\rm\hss \vskip \interauthorskip \@startauthor}
  \@startauthor \@author \@endauthor}
}}

\makeatother

\pdfoutput=1                    

\usepackage{amsmath,amsthm}
\usepackage{graphicx}
\usepackage{color}
\usepackage{amssymb,amsfonts,amsxtra,mathrsfs,bm}
\usepackage{multicol}
\usepackage{multirow}
\usepackage{paralist}
\usepackage{xspace}
\usepackage{algorithm}
\usepackage{algpseudocode}
\usepackage[colorlinks=true, urlcolor = blue]{hyperref}
\usepackage{bbm}
\usepackage{nicefrac}

\usepackage{subcaption}
\usepackage{caption}
\usepackage{float}
\usepackage{wrapfig}
\usepackage{enumitem}
\usepackage{mathtools}
\usepackage[capitalize,noabbrev]{cleveref}

\theoremstyle{definition}

\usepackage{booktabs} 
\usepackage{makecell} 

\numberwithin{equation}{section}

\title{Latent Structure of Affective Representations in Large Language Models}
\author{\name Benjamin J. Choi \email{benchoi@college.harvard.edu}\\
  \addr{Harvard University}
  \AND
  \name Melanie Weber \email{mweber@seas.harvard.edu}\\
  \addr{Harvard University}
}

\renewcommand{\mathbb}[1]{\ensuremath{\textup{\usefont{U}{msb}{m}{n}#1}}}

\hypersetup{
  colorlinks=true,
  linkcolor=blue,
  citecolor=blue,
  urlcolor=blue
}

\begin{document}
\maketitle

\begin{abstract}
The geometric structure of latent representations in large language models (LLMs) is an active area of research, driven in part by its implications for model transparency and AI safety. Existing literature has focused mainly on general geometric and topological properties of the learnt representations, but due to a lack of ground-truth latent geometry, validating the findings of such approaches is challenging. Emotion processing provides an intriguing testbed for probing representational geometry, as emotions exhibit both categorical organization and continuous affective dimensions, which are well-established in the psychology literature. Moreover, understanding such representations carries safety relevance. In this work, we investigate the latent structure of affective representations in LLMs using geometric data analysis tools. We present three main findings. First, we show that LLMs learn coherent latent representations of affective emotions that align with widely used valence--arousal models from psychology. Second, we find that these representations exhibit nonlinear geometric structure that can nonetheless be well-approximated linearly, providing empirical support for the linear representation hypothesis commonly assumed in model transparency methods. Third, we demonstrate that the learned latent representation space can be leveraged to quantify uncertainty in emotion processing tasks. Our findings suggest that LLMs acquire affective representations with geometric structure paralleling established models of human emotion, with practical implications for model interpretability and safety.
\end{abstract}

\section{Introduction}

Recent advances in mechanistic interpretability have illuminated how large language models (LLMs) internally represent high-level semantic features such as factual knowledge, syntax, and sentiment. A growing body of work investigates whether such features are geometrically organized in semantically plausible, useful ways \citep{skean2024does, tigges2023sentiment,lee2025shared}. 

Emotions provide a particularly compelling domain for studying geometric representations in LLMs because their semantic organization has been extensively studied in the psychology and neuroscience literature. Emotions can be described both in terms of discrete categories (e.g., anger, joy, fear) and along continuous affective dimensions such as valence (positive--negative) and arousal (calm--excited); in particular, this two-dimensional valence-arousal model (Fig.~\ref{fig:valence-arousal-comparison}) of emotion has proved particularly dominant in the psychology literature \citep{bradley1999affective, bliss2020immutability, maleki2023arps}. Because these structural models are based in human cognition, they provide a natural benchmark for probing whether LLMs encode emotional information in comparable ways to humans. If LLM latent spaces reproduce aspects of these psychological and neuroscientific models, it may not only deepen our understanding of how LLMs internally organize sentiment but also suggest potential parallels with natural intelligence. 

The present study is situated in an \emph{affective} context. More specifically, the affective computing literature has explored how computational systems can recognize and interpret human emotions across modalities such as speech and text \citep{calvo2010affect, huang2023emotionally, picard2000affective}. Inspired by this body of work, we design text-based emotion classification tasks to probe whether the LLM's learnt latent representations recover known categorical clusters; we hope to show, in line with previous ideas that distributed embeddings can encode affective dimensions \citep{shah2022affective}, whether continuous affective dimensions such as valence and arousal emerge as dominant axes of organization. In other words, rather than emphasizing the \emph{generation} of sentiment-bearing text, our analyses center on how models respond to (and internally structure) sentiment-encoded \emph{inputs}.

\paragraph{Related Work}

Our work engages with two prominent theories on the internal geometry of LLM representations. The first, the linear representation hypothesis, posits that high-level concepts are encoded as simple linear directions in activation space. This view is supported by studies showing that features like sentiment polarity can be identified with linear probes \citep{park2023linear} and manipulated through causal interventions along a linear axis to steer model outputs \citep{nanda2023emergent, park2023linear, jin2025exploring, tak2025mechanistic}. Conversely, the manifold hypothesis suggests that representations form more complex, nonlinear structures. For instance, recent analyses show that internal representations of semantic categories can form simplex-like hierarchical structures \citep{park2024geometry}, while other work has broadly identified geometric and manifold-like structures in neural representations of language \citep{mamou2020manifolds, he2024hierarchy}. By finding evidence for nonlinear latent structure in line with established psychological models, our work helps inform the current debate on representation geometry. 

Beyond representation geometry, a growing body of work has examined how LLMs process and represent emotion more broadly. \citet{huang2023emotionally} evaluated LLM emotional responses using standardized psychological instruments, while \citet{tak2025mechanistic} applied mechanistic interpretability tools to show that emotion representations are functionally localized within transformer models in a manner consistent with cognitive appraisal theory. Most directly related to our work, \citet{zhao2025emergence} analyzed probabilistic dependencies between emotional states in LLM outputs and found that models naturally form hierarchical emotion organizations aligning with Plutchik's emotion wheel, with larger models developing more complex hierarchies. While their analysis operates at the output-probability level rather than in internal activation space, their finding that structured emotion organization emerges in LLMs provides complementary motivation for our investigation of the geometry of these representations in the latent space.

Concurrent with our work, two independent studies have investigated emotion representations in LLMs through complementary lenses. \citet{sofroniew2026emotion} identify linear representations of emotion concepts in Claude Sonnet 4.5, showing that these representations are organized by valence and arousal and causally influence alignment-relevant behaviors such as reward hacking, blackmail, and sycophancy via activation steering. \citet{sun2026valence} decompose emotion steering vectors into a continuous two-dimensional valence--arousal subspace, demonstrating that steering along valence--arousal axes enables control over refusal and sycophancy. Both works corroborate our central finding that LLMs develop affective representations with geometric structure paralleling human psychological models; we discuss commonalities and differences in detail in Appendix~\ref{apdx:concurrent}.

\paragraph{Summary of Contributions}
The main contributions of our work are as follows:

\begin{enumerate}[leftmargin=*, itemsep=4pt, topsep=6pt, parsep=2pt]
    \item \textbf{Representational similarities with human affective models.} 
    We show that Gemma-2-9B, Mistral-7B, as well as LLaMA-3-70B-Instruct develop coherent internal representations of affective emotion. These representations align with established valence--arousal models from psychology.
    
    \item \textbf{Evidence for nonlinearity.}
    We find that the geometry of LLM emotion representations exhibits modest nonlinearity (in line with the parabolic curvature of valence-arousal space). While we find that affective emotion geometry is locally amenable to linear analysis, our evidence for nonlinear global structure suggests that a purely linear representation hypothesis is insufficient.
    
    \item \textbf{Applications to uncertainty quantification.} 
    We demonstrate that the geometry of these structured representation spaces can be exploited to quantify predictive uncertainty in emotion recognition tasks, illustrating both practical utility and interpretability gains.
\end{enumerate}

\noindent Additionally, in the appendices we provide complementary evidence from two directions: a parallel analysis showing that similar parabolic affective geometry emerges in human neural (EEG) data (Appendix~\ref{apdx:neural}), and causal steering experiments demonstrating that probe-derived emotion directions can be used as activation vectors to shift the emotional valence of generated text (Appendix~\ref{apdx:steering}).

\section{Background}

\paragraph{Constructing Latent Space Representations}
We will employ two manifold learning methods for recovering latent space representations of language model embeddings: classical multidimensional scaling (MDS) and Isometric Feature Mapping (Isomap). 

\emph{Classical MDS}~\citep{torgerson1952multidimensional}
takes a dissimilarity matrix $D=[d_{ij}]$ and constructs the doubly centered squared-distance matrix 
$B = -\tfrac{1}{2} J D^{\circ 2} J$ with $J = I - \tfrac{1}{n}\mathbf{1}\mathbf{1}^\top$. 
An eigendecomposition $B = Q \Lambda Q^\top$ then yields coordinates 
$z_i \in \mathbb{R}^k$ from the top-$k$ eigenpairs, chosen so that pairwise Euclidean distances in the embedding 
approximate the original dissimilarities in $D$. 
This closed-form solution provides a linear Euclidean representation of the data.

\emph{Isomap}~\citep{tenenbaum1997mapping} extends MDS to account for nonlinear structure by replacing the raw Euclidean distances $d_{ij}$ with estimates of geodesic distances along the data manifold. In practice, this is achieved by constructing a $k$-nearest neighbor ($k$NN) graph $G$ over the embeddings and computing approximate geodesic distances $\tilde{d}_{ij}$ as shortest-path distances on $G$; classical MDS is then applied to the geodesic distance matrix. By incorporating this local neighborhood structure, Isomap can capture curvature in the embedding space and reveal deviations from purely linear structure. 

\paragraph{Valence-Arousal Model of Emotion}

\begin{wrapfigure}{r}{0.42\linewidth}
    \vspace{-1.2em}
    \centering
    \includegraphics[width=\linewidth]{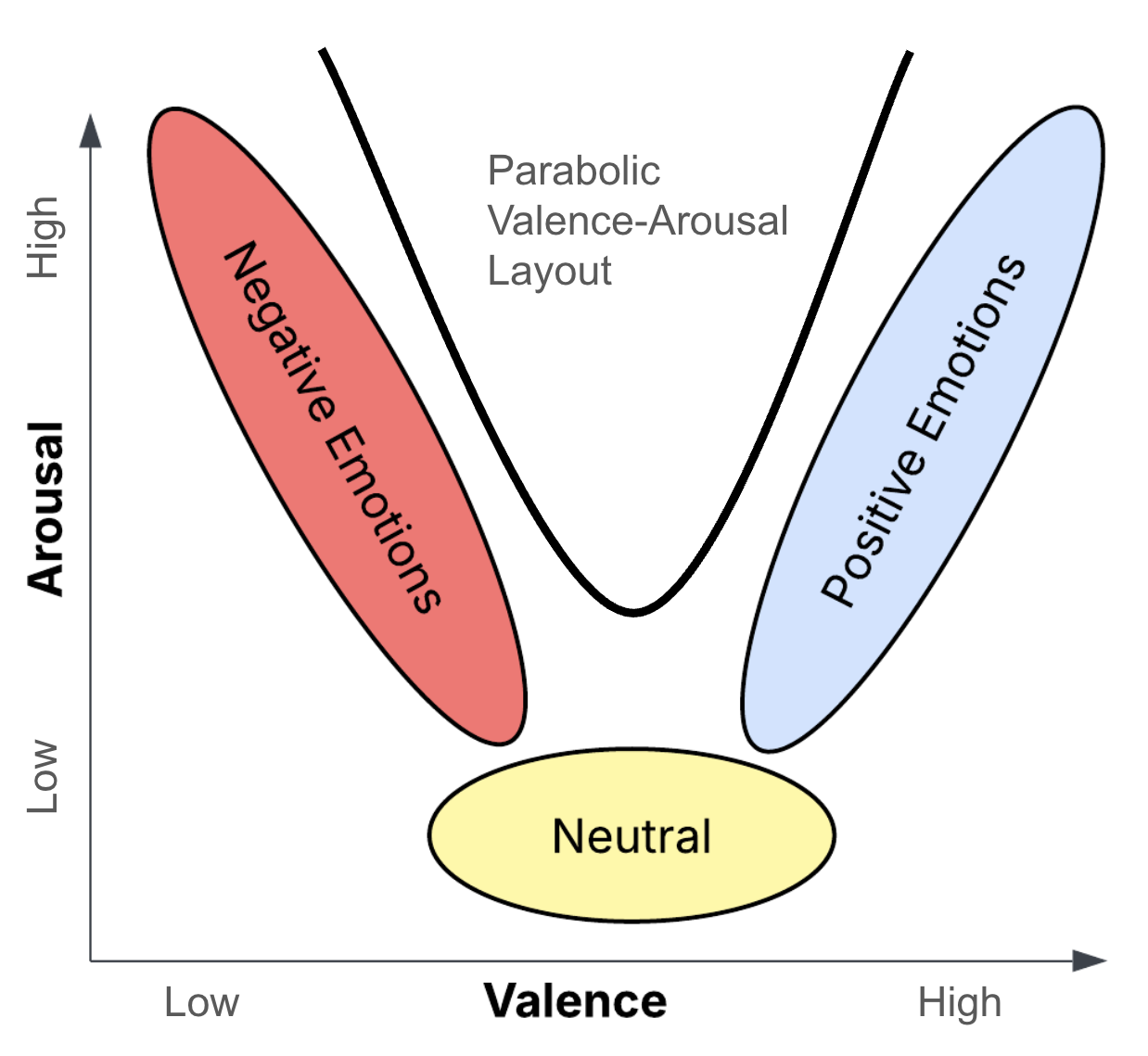}
    \caption{Parabolic valence--arousal model of affective space, based on \citet{maleki2023arps}.}
    \label{fig:valence-arousal-comparison}
    \vspace{-1.0em}
\end{wrapfigure}

A widely used framework in psychology and affective science conceptualizes emotions along two continuous dimensions: \emph{valence}, which captures the degree of pleasantness or unpleasantness, and \emph{arousal}, which indexes physiological activation or intensity \citep{russell1980circumplex, bliss2020immutability, kim2020situ, maleki2023arps}. Early work in this space originally posited circumplex emotion layouts centered around neutral \citep{russell1980circumplex}. In recent years, however, psychology literature has increasingly adopted the notion of a parabolic (i.e., ``V"-shaped) geometric layout of emotion \citep{kim2020situ, maleki2023arps}, owing to a general correlation between valence intensity and arousal arising in the typical distribution of human emotion. Geometric visualizations of the dual-axis valence-arousal emotion layout are shown in Figure~\ref{fig:valence-arousal-comparison}.

\section{Methods}

\subsection{Preliminaries}\label{sec:prelims}
To study the organization of emotional representations in LLMs, we begin by defining a notion of similarity between embeddings corresponding to different emotions. For each emotion pair $(i,j)$, we train a pairwise logistic regression classifier with L2 regularization on mean-pooled hidden state activations (see Section~\ref{sec:methodology}). We treat the test accuracy $\text{acc}_{ij}$ of this classifier as a proxy for the distance between emotions $i$ and $j$, with higher accuracy indicating greater separability. We then define by default $D_{ij} = \text{acc}_{ij}$ (all main findings hold under alternative metrics, including unsupervised cosine distance; see Appendix~\ref{apdx:cosine}). This yields a symmetric dissimilarity matrix that captures inter-emotion relationships in LLM activation space, since higher separation accuracy means the two emotions form well-separated activation clusters, while lower accuracy reflects overlapping representations. We also verified that our results are robust under alternative definitions of dissimilarity, such as activation cosine similarity (see Appendix~\ref{apdx:results}), distances to logistic regression hyperplanes, or alternative accuracy-to-distance transformations. Our logistic regression approach, however, is used by default as it provides a well-defined class-separating hyperplane, which plays a key role in our downstream uncertainty quantification application (see Section~\ref{sec:uncertainty}).

\subsection{Methodology}\label{sec:methodology}
The main aim of our exploratory analysis is investigating the following hypothesis:
\begin{quote}
    \emph{Do LLMs develop coherent internal representations of emotions that align with the valence-arousal model?}
\end{quote}
To test this hypothesis, we conduct an activation-based probing analysis across Mistral-7B and Gemma-2-9B centered around the geometry of representations in an explicit emotion-classification setting. In line with affective computing practices, we feed each prompt to the target LLM and prompt the model to classify the emotion from the randomized list of GoEmotions choices, subsequently storing activations corresponding to both correct and incorrect classifications. All inference was performed in zero-shot mode, with no additional fine-tuning on GoEmotions or related datasets.
For each text sample, we extracted mean-pooled hidden state activations from all transformer layers using forward hooks. Specifically, we collected the raw hidden states from each of the 32 layers in Mistral-7B and 42 layers in Gemma-2-9B, where each layer produces activations of dimension [1, sequence\_length, hidden\_size] since we process one sample at a time. We then applied mean pooling across the sequence dimension to obtain a single vector representation per layer for each model; we use layer outputs as opposed to sub-layer components, as in \citet{ju2024large} and \citet{li2024llms}. The activation vectors from correctly classified samples were used to train \emph{pairwise} classifiers between emotions, enabling us to identify how the geometric organization of emotional representations changes across depth and to assess where representations are most discriminative. The incorrect activations were stored for future downstream analysis (see Section~\ref{sec:uncertainty}). 

We focused on binary emotion comparisons rather than multiclass setups. This approach likewise avoids confounds from highly imbalanced class sizes (per-class correct sample counts range from 101 to 2,212 in Gemma-2-9B and 103 to 1,293 in Mistral-7B), yields clearer and more stable decision boundaries in activation space, and dovetails with MDS, which operates on pairwise dissimilarities. For each emotion pair,\footnote{To ensure sufficient dataset size, we conducted our analyses on emotion categories with at least 100 correct LLM classifications across the dataset, resulting in 20 emotions for Gemma-2-9B and 16 emotions for Mistral-7B (spanning positive, negative, ambiguous, and neutral valences).} we balanced the dataset by downsampling the majority class, then trained a logistic regression classifier with L2 regularization using an 80:20 train-test split. The resulting classification accuracies were subsequently converted into dissimilarity values $D_{ij}$ as described in Section~\ref{sec:prelims}, providing the pairwise distances for downstream MDS analysis.

\section{Exploratory Data Analysis}\label{sec:explo}

\subsection{Experimental Setting}

\paragraph{Models} We examine two state-of-the-art transformer-based LLMs: Gemma-2-9B~\citep{team2024gemma} and Mistral-7B~\citep{jiang2023mistral7b}. Gemma-2-9B is a 9-billion parameter, 42-layer model developed by Google DeepMind, while Mistral-7B is a 7-billion parameter, 32-layer model from Mistral AI. Although the exact training mixtures for these models have not been publicly disclosed, both were trained on large-scale, diverse web corpora that likely included sentiment-laden text, providing them with natural exposure to affective language. We also experimented with analyses on LLMs from the Qwen and LLaMA families but found that these models did not satisfy requisite sentiment recognition baselines necessary for our downstream experiments; further discussion is included in Appendix~\ref{apdx:details}.
In addition, to assess generalization across both scale and training regime, we conducted a full replication of our pipeline on LLaMA-3-70B-Instruct, a substantially larger and instruction-tuned model (see Section~\ref{sec:llama70b}).

\paragraph{Datasets}
Our primary dataset is \emph{GoEmotions}~\citep{demszky2020goemotions}, a manually annotated corpus of 58{,}009 English Reddit comments labeled for 27 fine-grained emotion categories plus \textit{Neutral}. The taxonomy comprises 12 positive, 11 negative, and 4 ambiguous categories, enabling analyses along both valence-aligned dimensions and discrete categories. Each comment was annotated by 3 or 5 independent raters (82 raters in total). We restrict our analyses to single-label examples exhibiting multi-rater agreement ($\sim$83\% of examples). As the largest manually annotated, fine-grained English emotion dataset to date, GoEmotions is a natural choice for our study.

\subsection{Semantic Analysis of Emotion Representations}
In our layer-by-layer activation analyses (see Appendix~\ref{apdx:results} for further details), we found that mean pairwise emotional separability generally rose toward the middle layers while exhibiting a slight drop at the final layers. Additionally, we found that Gemma-2-9B exhibited significantly higher activation separability compared to Mistral-7B, with mean test accuracies in the $>$0.9 range across all layers compared to $\sim$0.55 to $\sim$0.89 in Mistral-7B. This similarly aligns with Gemma-2-9B's greater LLM emotion recognition performance in the initial emotion classification setting, with an estimated $\sim$19.4\% correct classification rate in Gemma-2-9B compared to $\sim$13.7\% in Mistral-7B.

\begin{figure}[t]
    \centering
    \includegraphics[width=0.99\linewidth]{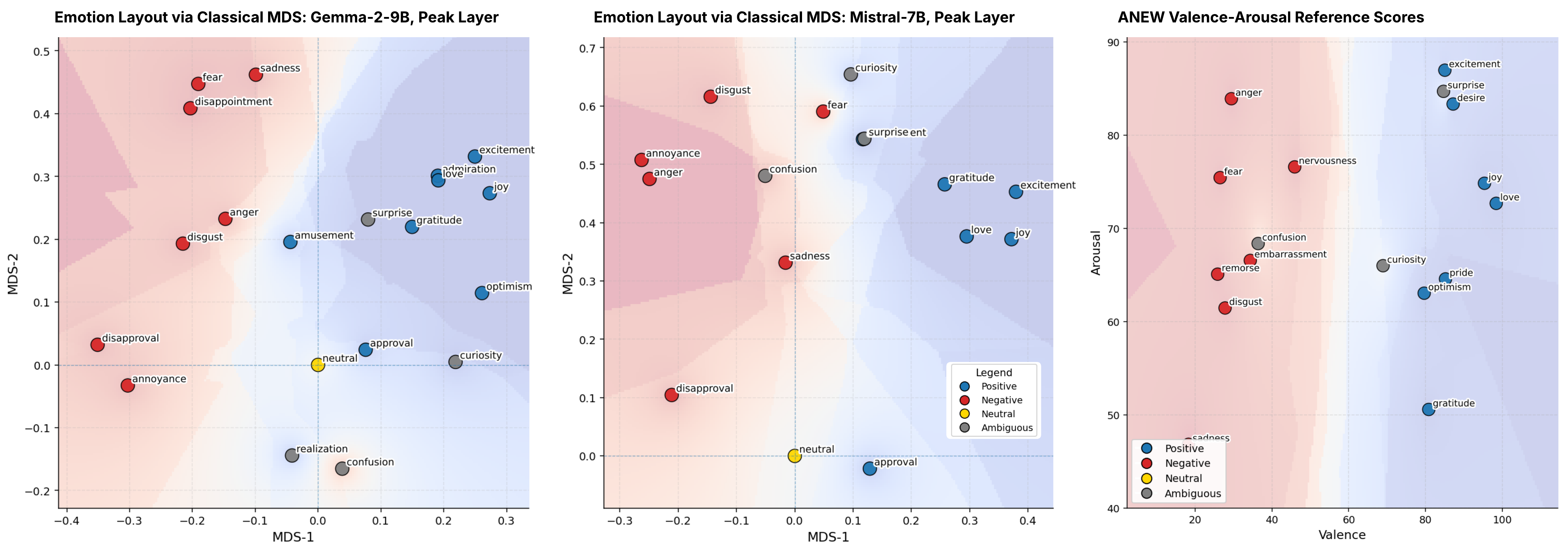}
    \caption{2D classical MDS embeddings of internal latent emotion representations in Gemma-2-9B (left) and Mistral-7B (middle), as well as ANEW reference valence-arousal scores (right). Layers chosen correspond to peak separability (Gemma-2-9B Layer 20, Mistral-7B Layer 27) and are representative of overall patterns observed throughout. Both plots are anchored to fixed orientations with neutral at the origin; colors correspond to the given GoEmotions taxonomy, with $k$NN background shading by valence included.}
    \label{fig:mds-results}
    \vspace{-0.5em}
\end{figure}

The latent representations obtained with classical MDS show that LLMs do indeed learn coherent representations that mirror logical semantic structure. Figure~\ref{fig:mds-results} shows the 2D projections of the classical MDS embeddings for Gemma-2-9B and Mistral-7B, respectively; both models exhibit clear semantic clustering, with positive emotions (joy, love, gratitude) clustering together in the upper-right quadrant and negative emotions (sadness, anger, disgust) grouping in the upper-left quadrant. 

\subsection{Comparison with Valence-Arousal Maps}\label{sec:valencearousal}

On a qualitative level, we observe notable similarities between our internal latent space representations and valence-arousal organization from the psychology literature. We find that the learned emotion structure (Figure~\ref{fig:mds-results}) across both LLMs appears to mimic the ``V"-shaped parabolic emotion layout, with neutral emotions positioned approximately at the vertex and positive (and negative) valences clustering together along a dual ``arm" structure. This semantic organization is visible in virtually all Gemma-2-9B layers and all but the early layers of Mistral-7B\footnote{These early Mistral-7B layers generally exhibit very low activation space emotion separability (see Appendix~\ref{apdx:results}), so the lack of semantic geometry here is consistent with the general observation that early Mistral-7B layers do not yet encode any coherent affective structure.}.

Beyond our qualitative support, we conduct a quantitative statistical test to compare our learned LLM representations against conventional valence-arousal maps. In particular, we compare the 2D MDS emotion embedding at each model layer with the map produced by third-party valence-arousal scores from the widely-used ANEW benchmark database from \citet{bradley1999affective}. The ANEW coordinates (see Figure~\ref{fig:mds-results}, right panel) for our statistical test come directly from human normative ratings, in which participants provided continuous valence and arousal scores for each lexical item. (We conduct our analyses using the 17 of the 28 classes in GoEmotions that are simultaneously covered in ANEW.) We then center both the MDS coordinates and the ANEW valence-arousal coordinates and align the MDS coordinates via scaled orthogonal Procrustes (single rotation and uniform scale), yielding fitted points corresponding to the MDS LLM latent representation. The test statistic is the Procrustes $R^{2}$, which measures the proportion of variance in the centered target configuration explained by the fitted configuration; significance is evaluated under a label-permutation null that shuffles the MDS emotion labels, recomputing the alignment and $R^{2}$ on each of $2{,}000$ permutations. We compute one-sided $p$-values of the observed $R^{2}$ relative to this null under emotion label permutation. (This procedure is invariant to arbitrary MDS orientation/scale and does not assume metric validity of the separability matrix; inference derives entirely from permutation.)

We find that 36 of 42 layers in Gemma-2-9B and 17 of 32 layers in Mistral-7B (including all of the final 14 layers in Mistral-7B) exhibit statistically significant alignment with the ANEW valence-arousal scores, providing additional quantitative evidence that the learned LLM latent representations exhibit semantically coherent structure.

\section{Evaluating Geometric Structure}\label{sec:geom}

Our analyses in the previous section are based on a linear embedding assumption via classical MDS. To further probe the intrinsic geometry of LLM emotion representations, we conduct two quantitative analyses. First, we assess the dimensionality of pairwise emotion distances by examining the eigenspectrum of the classical MDS Gram matrix. Second, we investigate the manifold hypothesis using Isomap \citep{tenenbaum1997mapping}, constructing a $k$NN graph, estimating geodesic distances, and embedding them with classical MDS. 

Eigenspectrum analyses (see Appendix~\ref{apdx:results}, Figure~\ref{fig:isomap_curvature}, left panel for an example), conducted across each individual layer of both Gemma-2-9B and Mistral-7B, reveal a generally diffuse rather than conclusively low-dimensional geometry. Under various monotone dissimilarity mappings ($D_{ij} = \text{acc}_{ij}$ and $D_{ij} = \max(0,2\cdot \text{acc}_{ij}-1)$), the participation ratio remains high (Gemma-2-9B $\sim\!17$, Mistral-7B $\sim\!9\text{--}14$), indicating that variance is spread across many eigenmodes\footnote{We note that as our dissimilarity metric is merely distance-like, we do observe occasional negative eigenvalues in our spectrum; however, only 1-2 negative eigendirections generally appear per layer, and these compose less than 1\% of overall variance mass across virtually all layers exhibiting emotional separability.}. This diffuse spectrum is consistent with what one would expect from a high-rank ``bulk'' component---akin to a Wigner-like distribution \citep{erdos2009local}---from noise and task-irrelevant variation, with meaningful structure appearing only as deviations from this bulk. Thus, our goal is not to claim a globally low-dimensional manifold, but rather to identify statistically significant axes that separate from the bulk and correspond to affective structure. We do so via a statistical eigengap test where we evaluate scale-invariant ratios $r_k=\lambda_k/\lambda_{k+1}$ between ordered eigenvalues against a permutation-generated null. This null distribution is generated by shuffling the off-diagonal of $D$ (symmetry preserved), and we flag eigenvalues where the observed ratio falls in the high tail ($p_{\mathrm{hi}}<0.05$). In general, our permutation test results (assessing prominence from the eigenspectrum bulk) appear generally consistent with a diffuse overall structure and no definitive fixed rank. We do find that Gemma-2-9B shows a recurring significant first eigengap with 16 out of 42 layers meeting a significance threshold at $k=1$, consistent with a dominant valence-like axis; we note, however, that this gap does not by itself establish any sort of fixed intrinsic dimension, but rather points to a recurring tendency toward a prominent individual axis of variation.

\begin{figure}[t]
    \centering
    \includegraphics[width=0.95\linewidth]{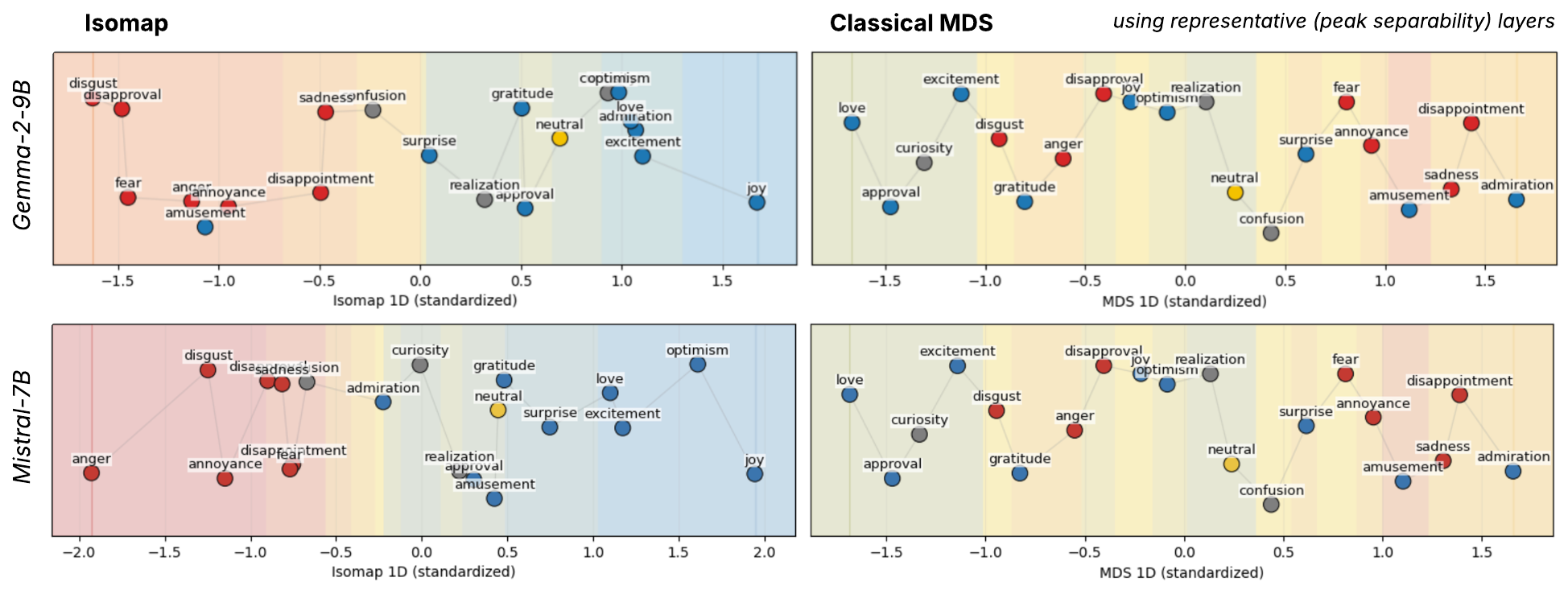}
    \caption{Isomap (\emph{left}) appears to unroll parabolic structure in the emotion data manifold, resulting in improvements in capturing rank-1 structure over classical MDS (\emph{right}). An artificial \emph{y}-axis jitter is introduced for label visibility, and we include $k$NN background shading by valence to illustrate improved semantic coherence (\emph{left}) under Isomap.}
    \label{fig:iclr7}
    \vspace{-0.5em}
\end{figure}

Our complementary Isomap evaluations allow us to probe for potential nonlinear manifold structure. For each layer, we constructed $k$NN\footnote{We select the Isomap $k$ parameter to maximize trustworthiness.} graphs over the pairwise separability matrix $|d'|$, computed geodesic distances, and compared Isomap against a classical MDS (Euclidean) baseline via two diagnostics: (i) relative trustworthiness (see Appendix~\ref{apdx:details} for formal definition) of local neighborhoods, and (ii) divergence between geodesic and Euclidean distances. On the overall trustworthiness front, we find generally minimal (i.e., $<0.03$) differences in trustworthiness between Isomap and classical MDS across ranks and choices of k--with the exception of the rank-1 embedding, for which Isomap demonstrates a median trustworthiness increase of 0.161 in Gemma-2-9B and 0.127 in Mistral-7B. We speculate that this rank-1 improvement is due to the aforementioned parabolic ``V"-shaped structure of the emotion data manifold, which Isomap naturally captures in ``unrolled" form in the 1D setting as shown in Figure~\ref{fig:iclr7}. 

In terms of discrepancies between geodesic (Isomap) vs. Euclidean (classical MDS) distances, we observe evidence consistent with modest nonlinearity. Specifically, distance ratios (i.e., geodesic / Euclidean) span from 1.00 to 1.80 from the tenth to ninetieth percentiles in Gemma-2-9B and 0.80 to 1.42 in Mistral-7B\footnote{We select the embedding dimension $d$ via an elbow-esque method where we identify the first residual variance drop less than 0.02.}. These results suggest that the LLM emotion spaces are almost Euclidean: high-rank, diffuse, and lacking strong low-dimensional curvature. The manifold nonlinearities that do appear in higher rank spaces appear consistent with the natural parabolic bending of valence--arousal space (e.g., see Appendix~\ref{apdx:results}, Figure~\ref{fig:isomap_curvature}, right panel, in addition to Figure~\ref{fig:iclr7}); the emotion data manifold exhibits a natural mode of curvature (due to the correlation between valence intensity and arousal) that manifests in LLM latent representations. On the whole, however, nonlinear manifold structure does not appear strong enough to render Euclidean representations ineffective as a basis for analysis.

\begin{wraptable}{r}{0.48\linewidth}
\vspace{-1.2em}
\caption{Mean trustworthiness improvement using Isomap over classical MDS per rank $d$}
\label{tab:trustworthiness}
\begin{center}
\begin{small}
\begin{sc}
\begin{tabular}{lcccc}
\toprule
Model & $d=1$ & $d=2$ & $d=3$ & $d=4$ \\
\midrule
Gemma   & 0.155 & -0.001 & -0.001 & 0.011 \\
Mistral & 0.099 & -0.024 & -0.026 & -0.011 \\
\midrule
 & $d=5$ & $d=6$ & $d=7$ & $d=8$ \\
\midrule
Gemma   & 0.013 & 0.020 & 0.024 & 0.022 \\
Mistral & -0.005 & 0.001 & -0.004 & -0.007 \\
\bottomrule
\end{tabular}
\end{sc}
\end{small}
\end{center}
\vspace{-1.0em}
\end{wraptable}

\section{Applications in Uncertainty Quantification}\label{sec:uncertainty}

During our pairwise logistic regression tests, we also saved activations corresponding to LLM misclassifications. Unlike the correct activation setting, where each activation is associated with a single emotion, each misclassification is linked to two emotions: the model outputted emotion, and the ground-truth GoEmotions label. Under a naive assumption of no semantic affective linkages, one might expect the misclassified activations to lie close in activation space to the ``correct'' activations for the LLM-outputted class. However, we instead observe a remarkable phenomenon where these misclassified activations instead lie in between the two emotions to which they are linked\footnote{Early layers lie slightly closer to the original prompt emotion, and late layers lie slightly closer to the erroneous outputted emotion (see Appendix~\ref{apdx:results}).}; that is, the ``misclassified'' activations hover much closer to the separating hyperplane between the two emotions (as defined by the corresponding pairwise logistic regression) than the ``correct'' activations do. We note that this phenomenon is not tautological, as we only evaluate distances to the separating hyperplane in terms of correct \emph{test} samples and misclassified activations, neither of which were included in the original logistic regression training data.

\begin{figure}[t]
    \centering
    \includegraphics[width=0.93\linewidth]{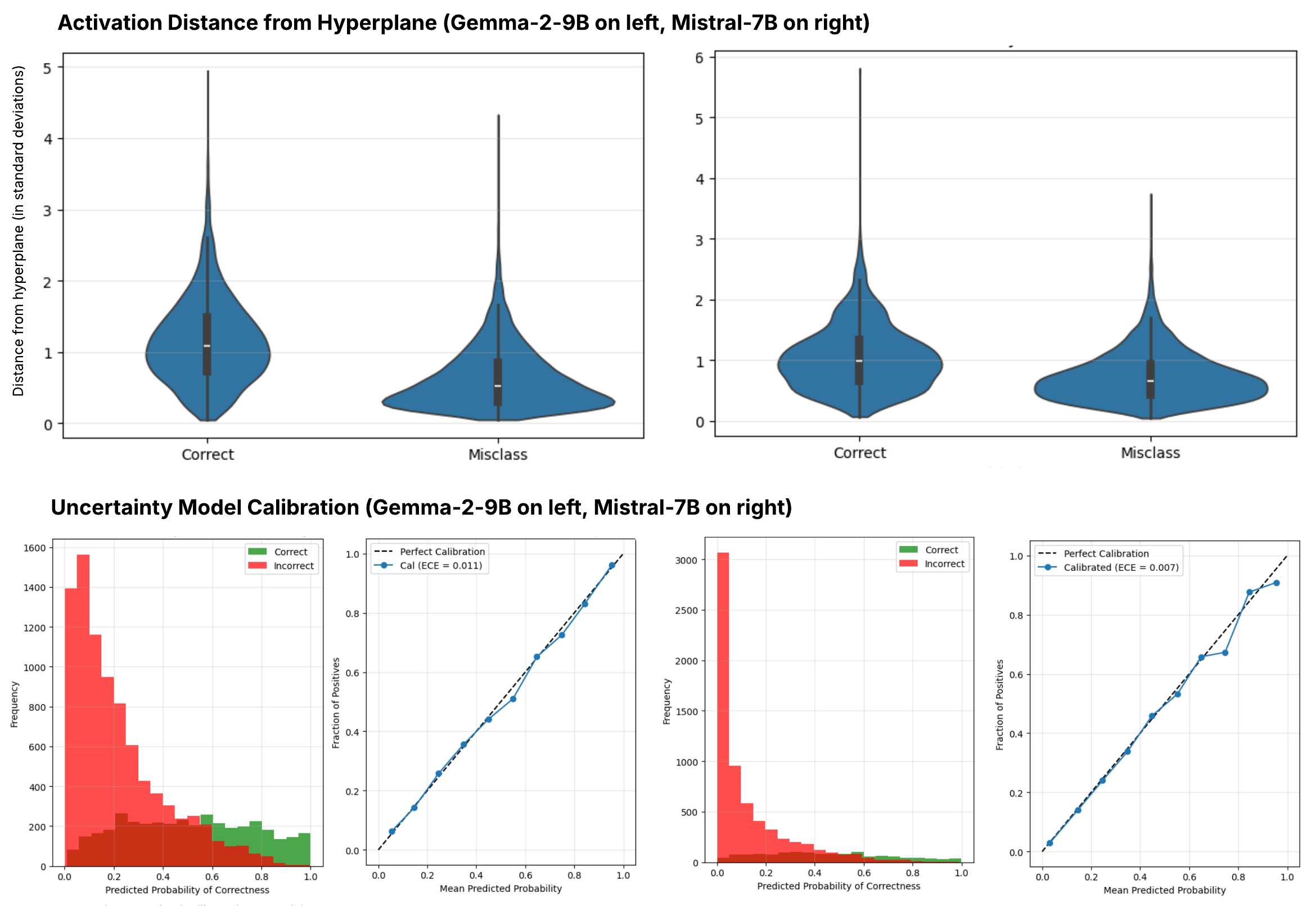}
    \caption{Upper panels depict activation distance from separating hyperplane, with misclassifications lying closer to the boundary than correct classifications. Lower panels depict test results of calibrated uncertainty quantification models, demonstrating robust calibration quality.}
    \label{fig:iclr10}
    \vspace{-0.5em}
\end{figure}

We can exploit this observation to generate practical utility by training a second round of pairwise logistic regression models on per-layer activation-to-hyperplane distances. Our work in this predictive \emph{uncertainty quantification} setting is motivated by the potential benefit of detecting LLM emotion processing misclassifications--the core concept, here, is to use the distance in activation space associated with an LLM prediction from the separating hyperplane between two classes as a measure of LLM ``confidence'' in that prediction. We restrict our analyses to emotion pairs with at least 25 correct and 25 misclassified samples in the data, resulting in 180 viable binary classification problems (3,858 correct samples vs. 8,748 misclassified) for Gemma-2-9B and 122 for Mistral-7B (5,254 correct vs. 15,216 misclassified) under a 60:20:20 train-val-test split. We use validation data in order to calibrate the logistic regression models, mapping raw hyperplane distances across all layers to well-calibrated probability estimates of correctness.

Our trained uncertainty models post 77.6\% accuracy (0.813 AUC-ROC) on Gemma-2-9B and 85.7\% accuracy (0.871 AUC-ROC) on Mistral-7B, compared to majority-class baselines of 69.4\% and 82.2\%, respectively. Importantly, post-calibration quality in terms of predicted correctness probability is strong: we find expected calibration errors of 0.011 (Gemma-2-9B) and 0.007 (Mistral-7B) on held-out test data. These results show that geometry-informed separating hyperplane distance--based regressions yield well-calibrated, discriminative uncertainty estimates in an LLM classification setting. 

\section{Generalization: LLaMA-3-70B-Instruct}
\label{sec:llama70b}

\begin{figure}[t]
    \centering
    \includegraphics[width=0.95\linewidth]{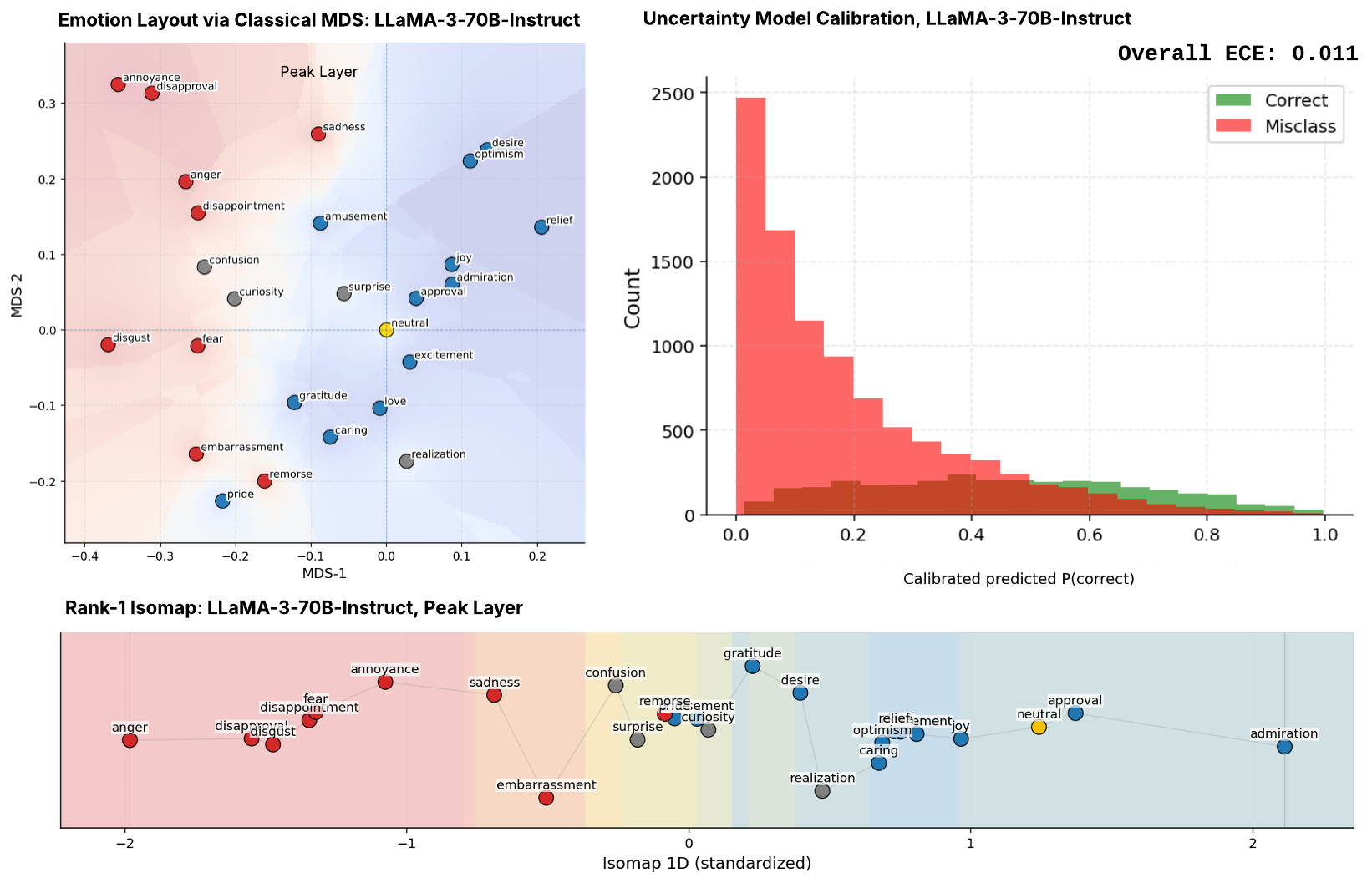}
    \caption{Corroboration of key results on LLaMA-3-70B-Instruct. Top-left: MDS emotion layout. Top-right: uncertainty model calibration. Bottom: Isomap embedding reflecting unrolled curvature.}
    \label{fig:llama70b}
    \vspace{-0.5em}
\end{figure}

To further assess the generalizability of our findings, we conducted a complete replication of our pipeline on LLaMA-3-70B-Instruct~\citep{llama3modelcard}, a substantially larger (70B parameter) and instruction-tuned model. Instruction tuning introduces supervised alignment and human-feedback--driven optimization, which can reshape representational structure; demonstrating stability of affective geometry under such changes therefore provides a more rigorous test of generalization.

LLaMA-3-70B-Instruct achieves a zero-shot emotion classification accuracy of 21.3\% on GoEmotions, outperforming both Mistral-7B (13.7\%) and Gemma-2-9B (19.4\%). Classical MDS embeddings reveal the same macroscopic affective layout observed in the earlier models, with positive and negative emotions diverging along two ``arms'' and neutral emotions near the geometric vertex (Figure~\ref{fig:llama70b}). The scaled orthogonal Procrustes test confirms alignment: 34 of 80 layers exhibit significant 2D alignment with $p_{2D}<0.05$, concentrated among later layers. As with the smaller models, Isomap recovers the expected parabolic structure in low-rank embeddings, and trustworthiness scores improve in the rank-1 setting relative to classical MDS. We also replicated the uncertainty quantification pipeline: misclassified samples again lie closer to pairwise separating hyperplanes than correct samples. On held-out test data, the uncertainty model posts 80.1$\%$ accuracy (0.822 AUC-ROC) with expected calibration error of 0.011 (baseline accuracy: 75.6\%). These results reinforce the generality of our findings across architectures, scales, and training paradigms; further details are in Appendix~\ref{apdx:results}, Figure~\ref{fig:llamafin}.

\section{Discussion}
Our analyses reveal that LLM emotion representations exhibit coherent geometric structure that aligns with affective models from psychology. Moreover, we show direct utility in the form of calibrated uncertainty models for LLM emotion processing, which leverage representation geometry to provide reliable estimates of predictive confidence.

\paragraph{Limitations}
First, it remains unclear whether the observed patterns generalize beyond the tested LLMs. Expanding our analysis to a wider range of models and families is an important direction for future work, as is a systematic study of the effects of emotion-focused fine-tuning. 
Second, our use of pairwise logistic regressions in conjunction with MDS and Isomap may not fully capture potential geometric intricacies; distributional analyses and advanced topological tools could reveal nuanced relationships in future work. Third, our primary GoEmotions dataset could have cultural or linguistic biases that constrain generality. Extending the analysis to more naturalistic prompts and to sub-layer components (e.g., attention vs.\ MLP) is also a valuable direction for future work.

\paragraph{Broader Impacts}
The parallels we uncover between LLM and human affective representations highlight potential similarities between artificial and human cognition (see Appendix \ref{apdx:neural}), suggesting that geometric priors rooted in human psychology may inform future interpretability frameworks. Our uncertainty quantification method also illustrates a principled way to detect likely misclassifications, with relevance to selective prediction, human-in-the-loop systems, hallucination detection, and other safety-critical applications. Additionally, our supplemental steering experiments (Appendix~\ref{apdx:steering}) 
provide causal evidence that these geometric directions are not merely 
diagnostic but functionally involved in the model's generation process, 
with implications for fine-grained behavioral control of LLM outputs.

\paragraph{Reproducibility Statement}
To facilitate replication, we provide anonymized code to reproduce experiments, figures, and statistical tests at \href{https://drive.google.com/file/d/1BWcCEOftiBmPMjjL9V61F0yaz5-ZXaff/view?usp=sharing}{this link}. Further experimental details are in Appendix~\ref{apdx:details}.

\paragraph{Acknowledgements}
MW acknowledges support from NSF award DMS-2406905 and Coefficient Giving.

\newpage
\bibliographystyle{plainnat}
\bibliography{ref}

\clearpage
\appendix

\section{Additional Experimental Results}\label{apdx:results}

\begin{figure}[h]
    \centering
    \includegraphics[width=0.9\linewidth]{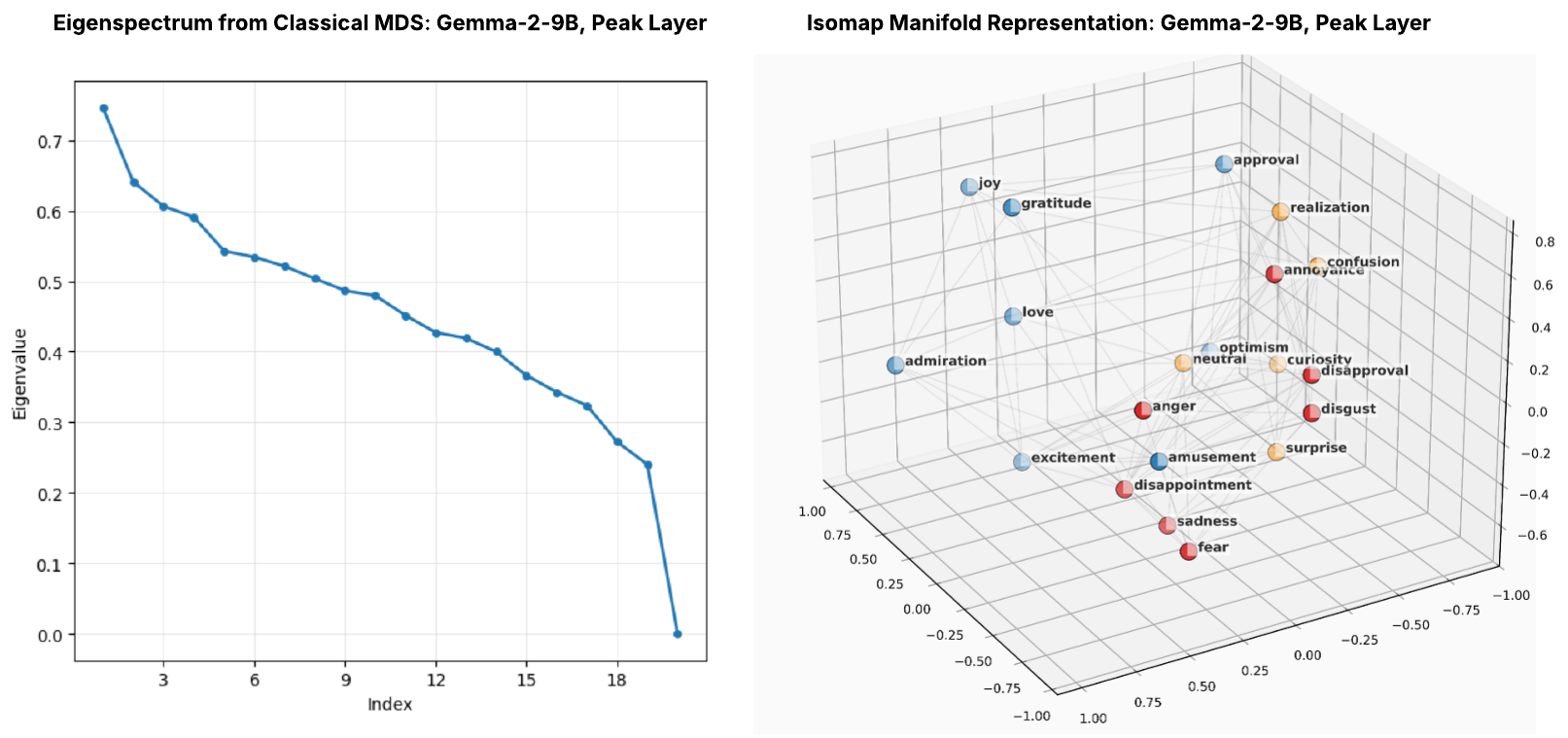}
    \caption{Representative examples of a classical MDS eigenspectrum (\emph{left}) and an Isomap embedding visualization (\emph{right}).}
    \label{fig:isomap_curvature}
\end{figure}

\begin{figure}[h]
    \centering
    \includegraphics[width=0.95\linewidth]{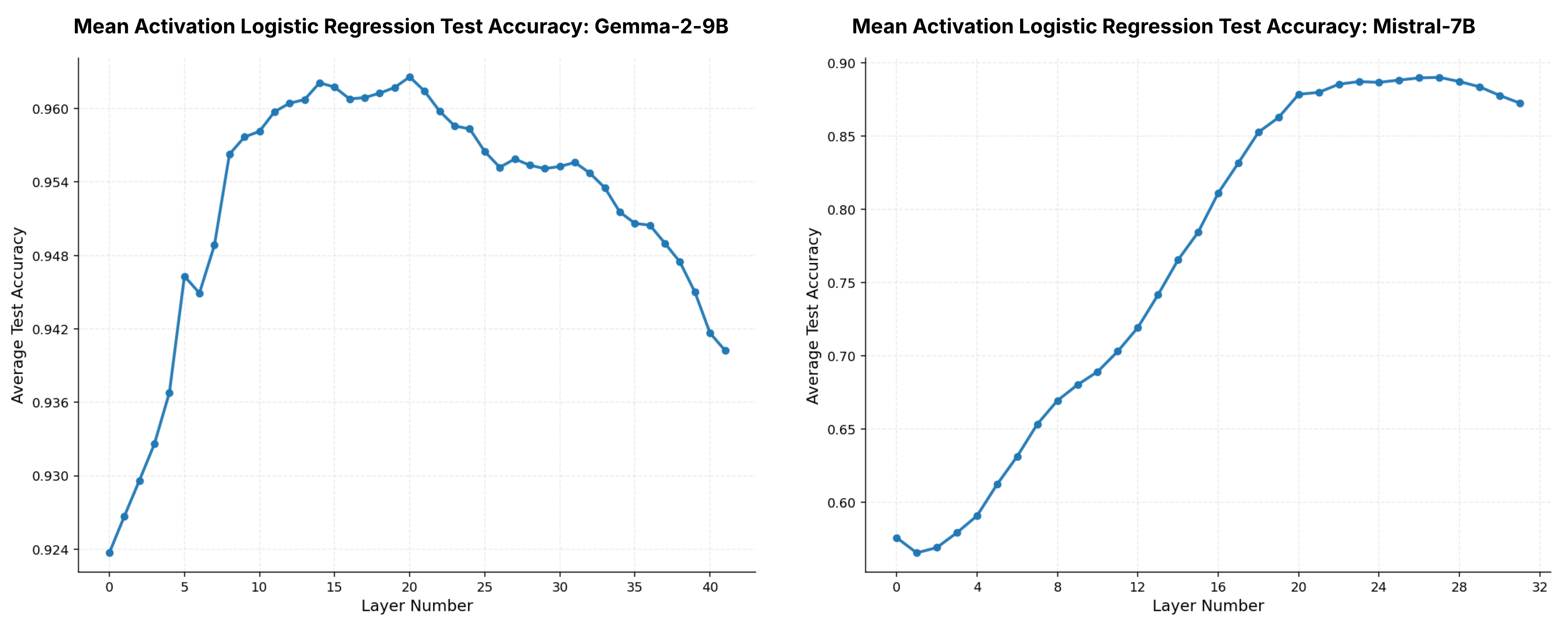}
    \caption{Mean emotional separability across layers for Gemma-2-9B (left) and Mistral-7B (right). Separability increases with depth before tapering off in the final layers, with Gemma-2-9B exhibiting higher separability than Mistral-7B.}
    \label{fig:layer-accuracy}
\end{figure}

\begin{figure}[h]
    \centering
    \includegraphics[width=0.9\linewidth]{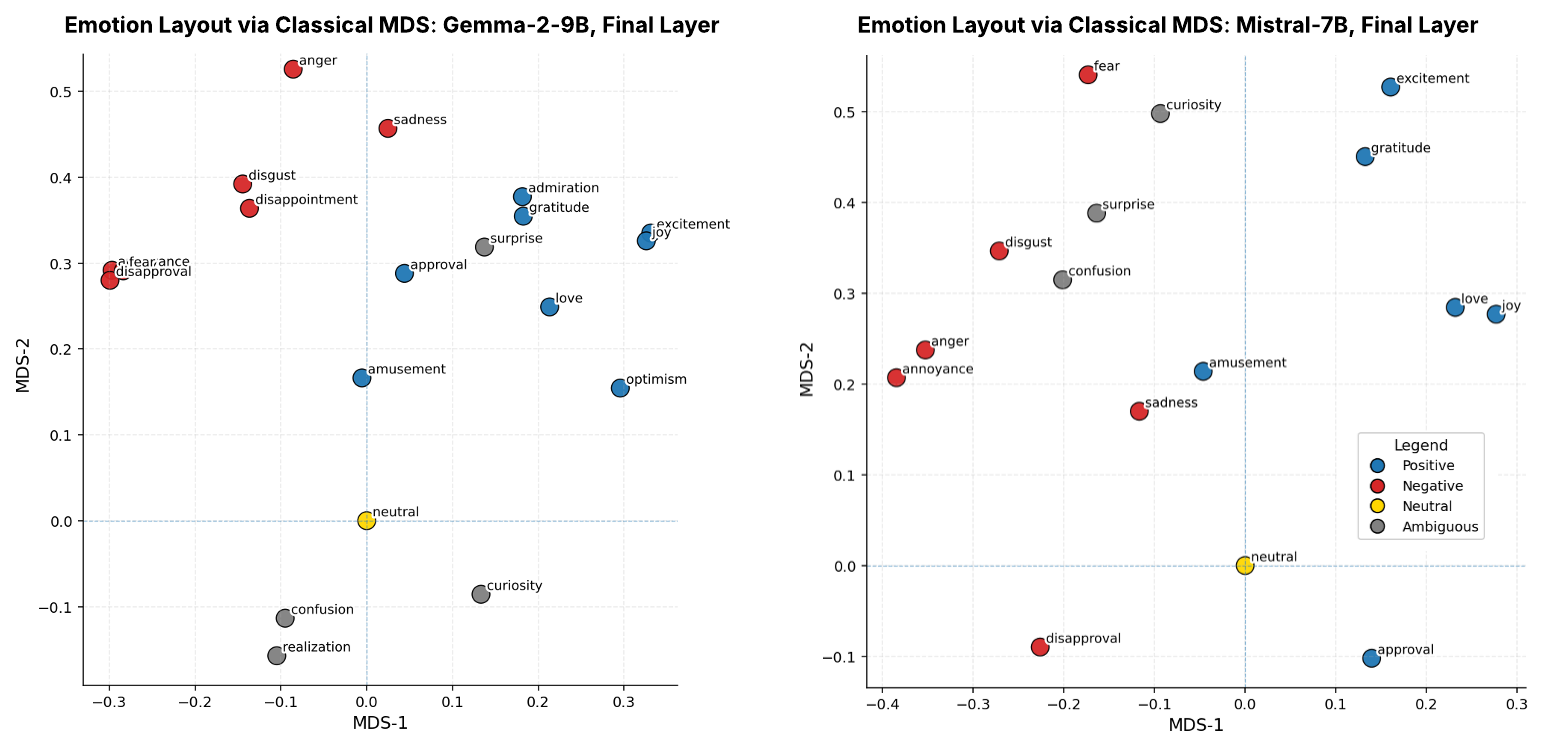}
    \caption{MDS emotion layouts for final layers in Gemma-2-9B (\emph{left}) and Mistral-7B (\emph{right}).}
    \label{fig:layer-accuracy3}
\end{figure}

\begin{figure}[h]
    \centering
    \includegraphics[width=\linewidth]{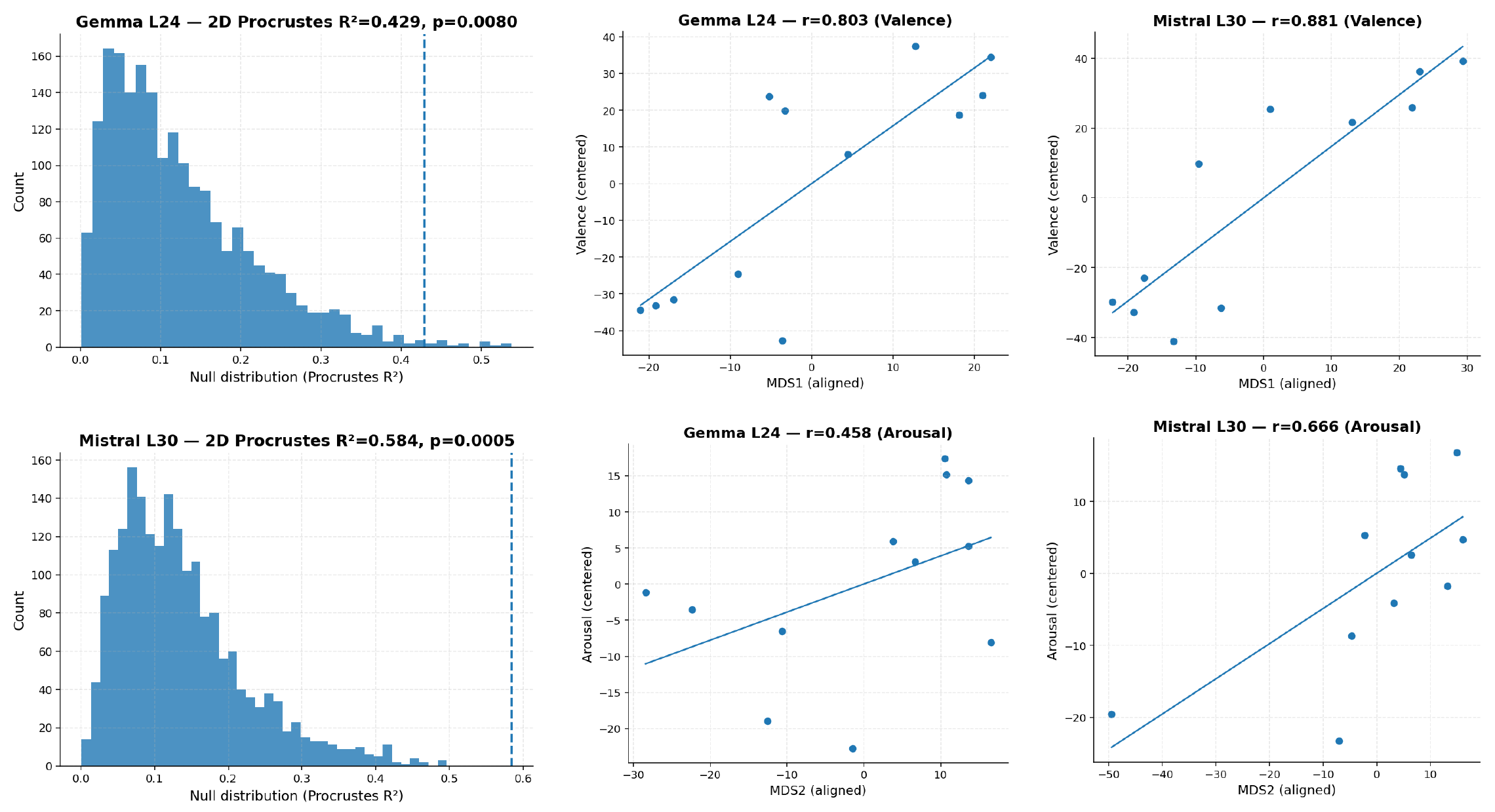}
    \caption{$R^2$ and $p$-values for the classical MDS LLM latent representations vs. established valence-arousal scores.}
    \label{fig:layer-accuracy4}
\end{figure}

Figure~\ref{fig:isomap_curvature} depicts further results from our analyses in Section~\ref{sec:geom}. Figures~\ref{fig:layer-accuracy}, \ref{fig:layer-accuracy3}, and \ref{fig:layer-accuracy4} depict further results from our pairwise logistic regression and classical MDS experiments. The mean emotional separability between discrete emotion classes for each layer in Figure~\ref{fig:layer-accuracy} is measured via pairwise logistic regression test accuracies.

\begin{figure}[h]
    \centering
    \includegraphics[width=\linewidth]{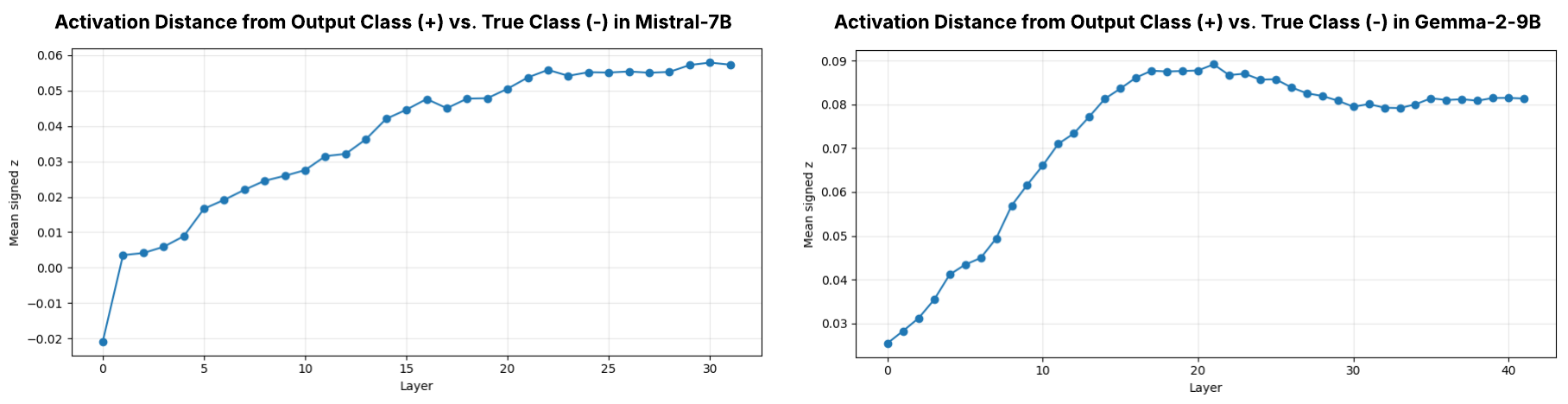}
    \caption{Layer-by-layer activation distances from output class vs. ground-truth input class. Higher values signify relatively closer distances to the model output emotion class.}
    \label{fig:layer-accuracy2}
\end{figure}

\begin{figure}[t]
    \centering
    \includegraphics[width=0.9\linewidth]{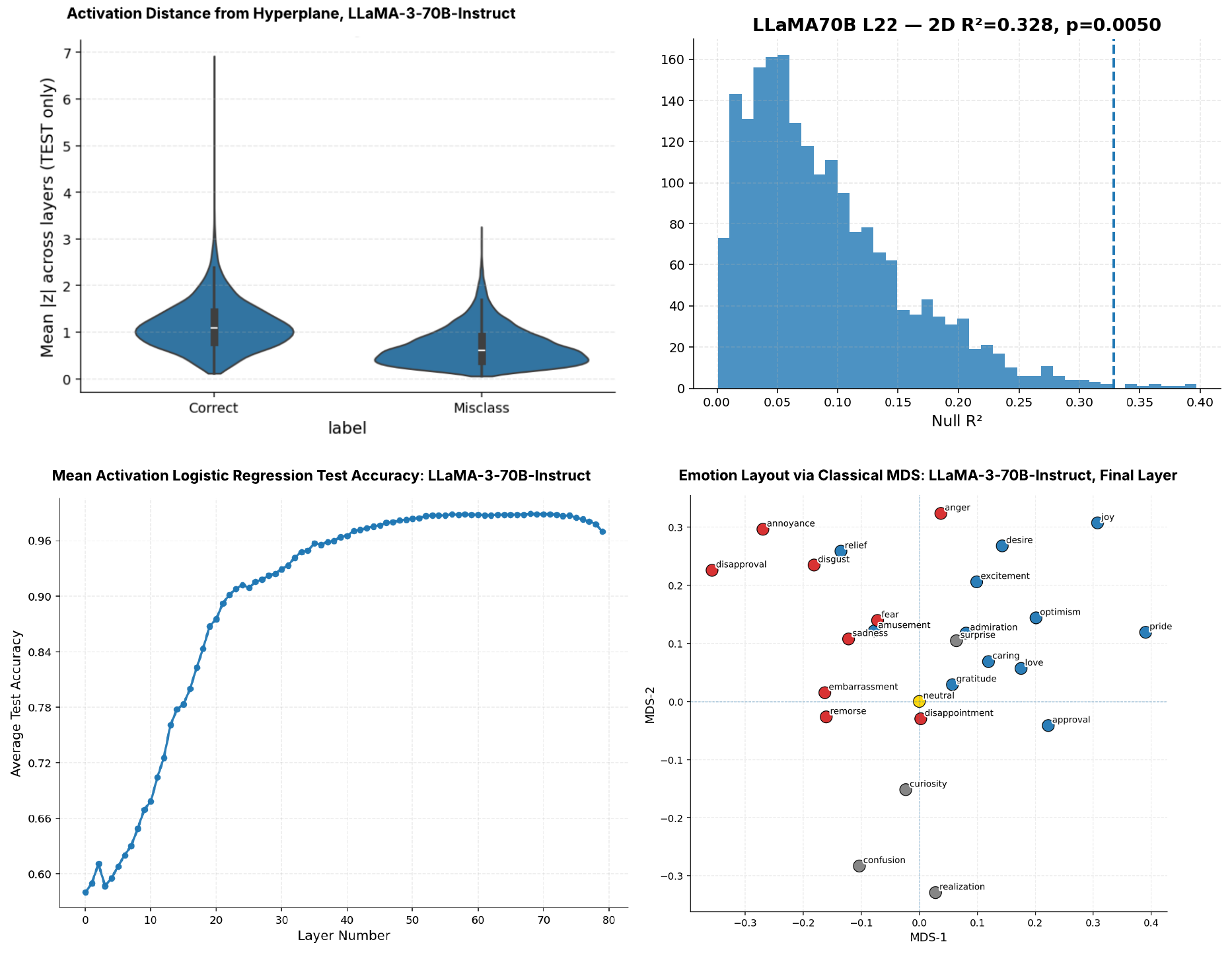}
    \caption{Additional results from experiments on LLaMA-3-70B-Instruct. Top-left: activation distance from hyperplane for correct vs. misclassified samples. Top-right: statistical significance result for ANEW alignment. Bottom-left: mean logistic regression accuracy by layer. Bottom-right: classical MDS emotion layout for final layer.}
    \label{fig:llamafin}
\end{figure}

Figure~\ref{fig:layer-accuracy4} depicts results from our statistical alignment test between the classical MDS embeddings and established valence-arousal scores, showing results ($R^2$ and $p$-values) for two statistically significant layers in Gemma-2-9B and Mistral-7B. Figure~\ref{fig:layer-accuracy2} depicts layer-by-layer trends from our pairwise logistic regressions on the out-of-sample misclassified pairs. More specifically, lower (or more negative) y-axis values signify that activations are closer to the ground truth input emotion, while higher values signify that activations are closer to the output emotion. Misclassified activations, as discussed in Section~\ref{sec:uncertainty}, lie closer to the separating hyperplane than correct activations; as shown in Figure~\ref{fig:layer-accuracy2}, activations trend away from the ground truth input emotion and toward the model output emotion as model layers progress. Figure~\ref{fig:llamafin} depicts additional results from our experiment generalizing our findings to LLaMA-3-70B-Instruct.

\begin{figure}[h]
    \centering
    \includegraphics[width=\linewidth]{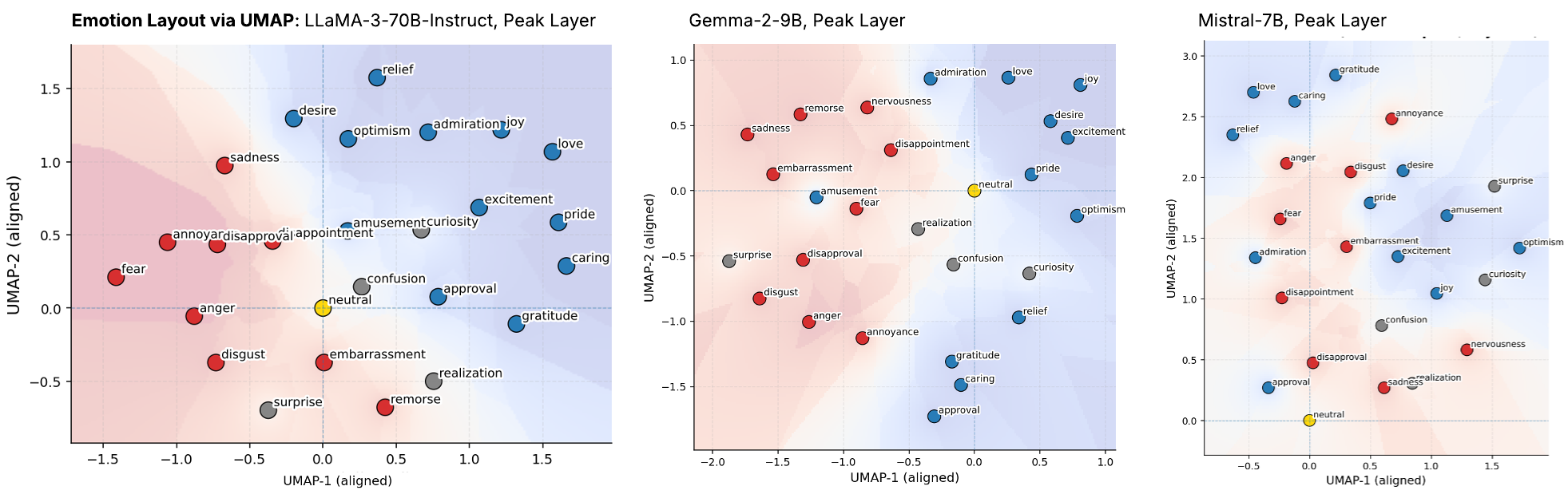}
    \caption{UMAP visualizations for Gemma-2-9B, Mistral-7B, and LLaMA-3-70B-Instruct.}
    \label{fig:umapnew}
\end{figure}

\subsection{UMAP Visualizations of Distance Geometry}
\label{apdx:umap}

To complement our classical MDS and Isomap analyses, we additionally conducted a supplementary analysis using UMAP~\citep{mcinnes2018umap}. For each model (Gemma-2-9B, Mistral-7B, and LLaMA-3-70B-Instruct) and each layer, we performed a small UMAP hyperparameter sweep over the number of neighbors $n_{\text{neighbors}} \in \{3,\dots,12\}$ as opposed to proceeding directly to classical MDS. For each setting we computed the trustworthiness~\citep{venna2001neighborhood} of the resulting embedding with respect to the original distance matrix, and selected the $n_{\text{neighbors}}$ that maximized trustworthiness. Visualizations of the 2D UMAP case for representative layers are shown in Figure~\ref{fig:umapnew}. We also verified the presence of statistically significant alignment patterns between these UMAP embeddings and the ANEW reference scores, with 34 significant layers in LLaMA-3-70B-Instruct, 10 in Mistral, and 41 in Gemma (p $<$ 0.05).

Across these settings, UMAP recovers general semantic clustering classical MDS, with positive, neutral, and negative sentiments exhibiting relatively coherent and clustered layouts. These results indicate that our observed semantic structure is stable under a nonlinear, neighborhood-preserving embedding applied directly to the pairwise distance geometry.

\subsection{Cosine-Distance Experiment}
\label{apdx:cosine}

Our main analyses define dissimilarities via pairwise logistic-regression separability, which couples the inferred geometry to a particular supervised probing pipeline. To test whether our conclusions depend critically on this choice, we performed an additional ablation based on a purely geometric cosine-distance metric over the mean-pooled activations.

\begin{figure}[h]
    \centering
    \includegraphics[width=0.45\linewidth]{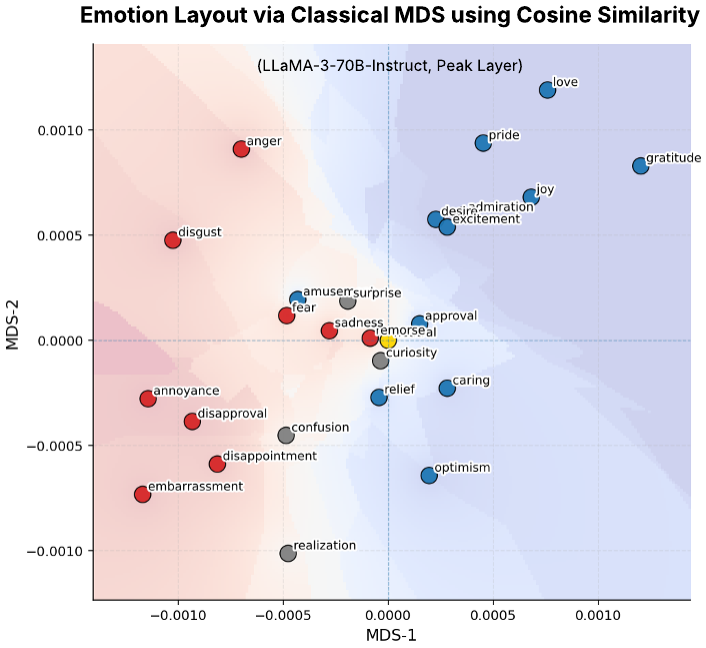}
    \caption{Classical MDS embeddings obtained from cosine-based dissimilarities between mean activation vectors.}
    \label{fig:cosine}
\end{figure}

Concretely, for emotion class across layers, we first computed the mean activation vector by averaging the mean-pooled hidden states over all correctly classified examples of that emotion (as in the main pipeline). We then constructed an alternative distance matrix
\[
D^{\text{cos}}_{ij} \;=\; 1 - \cos\bigl(\bar{h}_i, \bar{h}_j\bigr),
\]
where $\bar{h}_i$ and $\bar{h}_j$ denote the mean activation vectors for emotions $i$ and $j$, and $\cos(\cdot,\cdot)$ is the cosine similarity. As before, diagonals were set to zero, missing entries were imputed by the global mean, and the matrix was symmetrized.

Using $D^{\text{cos}}$, we then repeated the classical MDS procedure; Figure~\ref{fig:cosine} shows an example of these cosine-based embeddings. Qualitatively, we again observe clear semantic clustering; quantitatively, recomputing our Procrustes $R^2$ ANEW alignment test, we also observe broad statistically significant alignment across all layers. Taken together, this cosine-distance experiment helps support the robustness of our main findings by corroborating our affective representation geometry.

\section{Comparison with Neural Data}\label{apdx:neural}

To further strengthen our claim of similar structure between human affective processing and LLM internal representations, we conducted an additional parallel analysis exploring whether similar semantic representations of emotion emerge in human brainwave data. Using affective emotional data from 123 human subjects across 32 EEG channels (28 individual clips covering nine distinct emotions) from the FACED dataset \citep{chen2023large}, we investigate latent structure in the internal emotional landscape present in human neural data. We employed an experimental setting designed to resemble our LLM analysis: we trained pairwise logistic regressions on emotional clip classification and translated statistically significant test discriminatory accuracies into neural distances. (Statistical significance was computed via label permutation with a 0.05 p-value cutoff to account for the inherent noise present in neural data.) To address the high dimensionality of human brainwave data, we binned the neural signals into mean and variance summary statistics corresponding to conventional neural frequency bands (delta, theta, alpha, beta, and gamma) before performing pairwise logistic regressions. The resulting distances informed a downstream classical MDS analysis depicting the latent geometry of emotions in affective EEG data.

\begin{figure}[h]
    \centering
    \includegraphics[width=0.93\linewidth]{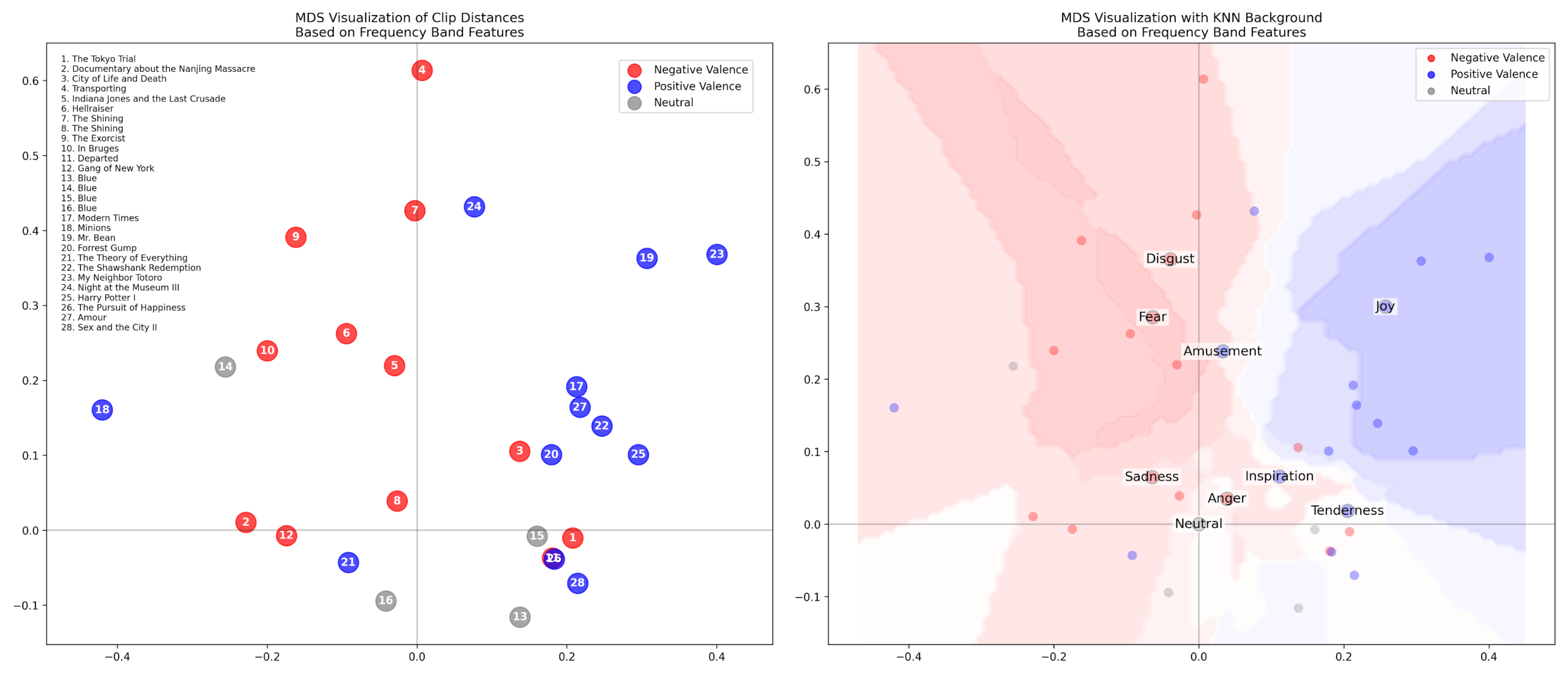}
    \caption{We find that a similar parabolic ``V"-shaped emotion layout exists in human brainwave data, corroborating a link between human cognition and LLM affective emotion processing. Left panel depicts individual affective samples colored by valence; right panel depicts center of mass points for each emotion label alongside $k$NN shading by valence.}
    \label{fig:iclr4}
\end{figure}

As shown in Figure~\ref{fig:iclr4}, the embedding space exhibits a parabolic, ``V"-shaped configuration. Neutral emotions are positioned near the vertex of the structure, while emotions with positive and negative valence again diverge along two distinct arms. This parallel suggests that the structural patterns observed in LLM representations may reflect principles of emotional representation that also characterize human neural activity, offering additional empirical support for the link between human affective cognition and learned representations in LLMs.

\paragraph{FACED Dataset}
As a secondary dataset to probe parallels between text-derived and neural emotion representations, we use \emph{FACED} \citep{chen2023large}, a large, open EEG resource with recordings from 123 participants using 32 electrodes (10--20 system). Participants watched 28 emotion-eliciting video clips spanning nine categories---four positive (amusement, inspiration, joy, tenderness), four negative (anger, fear, disgust, sadness), and one neutral. We use FACED to test whether the geometric structure we observe in LLM embeddings also exhibits potential similarities with structure seen in neural EEG responses, thereby further investigating a potential link between affective artificial and natural cognition.

\paragraph{Limitations of Neural Data Analysis}
Our secondary analysis regarding similar patterns in human brainwave data exhibits several limitations that we hope to address in future work. While our results present first evidence for the alignment of latent structure in LLM representations and human brainwave data, this does not constitute evidence of shared cognitive encoding. Furthermore, our observations are based on a single dataset; a study with larger scope is an interesting avenue for future work. An analysis of the impact of cultural biases in the dataset was likewise beyond the scope of the present paper.

\section{Additional Experimental Details}\label{apdx:details}

\paragraph{GoEmotions Data} The GoEmotions corpus used in this study contains the following emotion categories: admiration, amusement, anger, annoyance, approval, caring, confusion, curiosity, desire, disappointment, disapproval, disgust, embarrassment, excitement, fear, gratitude, grief, joy, love, nervousness, optimism, pride, realization, relief, remorse, sadness, surprise, and neutral. 

For Mistral, the correct sample data distribution included 47 samples of admiration, 1293 of amusement, 509 of anger, 228 of annoyance, 275 of approval, 31 of caring, 123 of confusion, 312 of curiosity, 29 of desire, 93 of disappointment, 373 of disapproval, 213 of disgust, 86 of embarrassment, 218 of excitement, 201 of fear, 670 of gratitude, 203 of joy, 1048 of love, 35 of nervousness, 60 of optimism, 36 of pride, 89 of realization, 40 of relief, 744 of sadness, 291 of surprise, and 103 of neutral. 

For Gemma, the correct sample counts were 742 for admiration, 1211 for amusement, 455 for anger, 411 for annoyance, 216 for approval, 95 for caring, 481 for confusion, 675 for curiosity, 88 for desire, 439 for disappointment, 1148 for disapproval, 177 for disgust, 60 for embarrassment, 304 for excitement, 154 for fear, 678 for gratitude, 284 for joy, 300 for love, 37 for nervousness, 208 for optimism, 36 for pride, 101 for realization, 55 for relief, 58 for remorse, 235 for sadness, 318 for surprise, and 2212 for neutral.

For LLaMA-3-70B-Instruct, the correct-sample distribution included 1105 samples of admiration, 1555 of amusement, 472 of anger, 222 of annoyance, 518 of approval, 246 of caring, 407 of confusion, 689 of curiosity, 67 of desire, 265 of disappointment, 1225 of disapproval, 240 of disgust, 70 of embarrassment, 249 of excitement, 182 of fear, 1135 of gratitude, 223 of joy, 341 of love, 203 of optimism, 47 of pride, 84 of realization, 85 of relief, 103 of remorse, 333 of sadness, 307 of surprise, and 1233 of neutral.

As mentioned in Section~\ref{sec:explo}, we also conducted exploratory experiments with Qwen and LLaMA models. Specifically, we ran our affective classification workflow on \href{https://huggingface.co/Qwen/Qwen2.5-7B}{Qwen2.5-7B} and \href{https://huggingface.co/meta-llama/Llama-2-7b}{LLaMA-2-7B}, but found that neither model produced sufficient correct classifications (i.e., $>$100 across more than three individual categories) on the GoEmotions dataset which are requisite for our downstream analyses.

\paragraph{Classification Routine} We analyze Gemma-2-9B, Mistral-7B, and LLaMA-3-70B-Instruct using AutoModelForCausalLM at float16 precision. For single-label predictions, each classification prompt shuffles the vocabulary of emotions to reduce positional bias. The prompt template is: ``Classify this text into exactly one emotion from this list: \ldots\ Text: \{text\} Emotion:''. The first token generated after the word ``Emotion:'' is decoded and matched to the enumerated set, with unmatched predictions skipped from activation saving. To capture internal activations, we register forward hooks on every transformer block. The hooks extract hidden states, apply mean pooling across the sequence dimension to produce a single vector per layer, and store outputs as detached tensors. The shuffled emotion order is reused to maintain consistency between prediction and activation passes. 

Balanced pairwise training is carried out by constructing equal-sized datasets for every emotion pair $(e_i, e_j)$. Concatenated data matrices $\mathbf{X} \in \mathbb{R}^{2m \times H}$ with binary labels $\mathbf{y} \in \{0,1\}$ are split in an 80/20 ratio with stratification. Logistic regression models with maximum iterations of 1000 are trained and evaluated, and we record training and test accuracy, decision margins, per-point correctness, and sample counts to assist in downstream analyses.

\paragraph{Statistical Alignment Test} As discussed in Section~\ref{sec:valencearousal}, to align model-derived spaces with our external valence-arousal ratings for our statistical alignment test, we use an orthogonal Procrustes transformation. Let $Y \in \mathbb{R}^{n \times 2}$ denote the valence-arousal matrix, centered column-wise, and let $X$ denote the corresponding model embedding. The alignment is defined as
\[
\min_{R \in \mathbb{R}^{2\times 2}, R^\top R = I,\, a \in \mathbb{R}} \;\|a X_c R - Y_c\|_F^2,
\]
where $X_c$ and $Y_c$ are row-centered. The solution is obtained by singular value decomposition $X_c^\top Y_c = U\Sigma V^\top$, with $R=UV^\top$ and $a = \mathrm{tr}(\Sigma)/\|X_c\|_F^2$. The alignment coefficient of determination is reported as
\[
R^2_{\text{proc}} = 1 - \frac{\|aX_cR - Y_c\|_F^2}{\|Y_c\|_F^2}.
\]

Significance of alignment is assessed via permutation tests. Row labels of $Y$ are permuted and $R^2_{\text{proc}}$ is recomputed for $T=2000$ shuffles. With observed value $R^2_{\text{obs}}$, the $p$-value is
\[
p = \frac{1 + \sum_{t=1}^{T} \bm{1}\{R^2_{(t)} \ge R^2_{\text{obs}}\}}{T+1},
\]
with reproducibility guaranteed via fixed random seed. After alignment, axis-wise correlations are calculated between aligned $X_{\text{hat}}$ and $Y_c$, with Pearson correlations $r_{\text{val}}$ for valence and $r_{\text{aro}}$ for arousal; these metrics are shown in Figure~\ref{fig:layer-accuracy4}. These correlations are also subjected to permutation tests that re-align for each shuffle, ensuring axis assignment does not bias significance.

Practical safeguards include mean pooling across sequence length to prevent padding sensitivity, imputation of missing similarity values by the global mean to avoid distortions, clamping of negative eigenvalues from double-centering to zero for stability, and centering of both $X$ and $Y$ prior to Procrustes analysis so that explained variance is measured relative to mean-centered data. Orthogonal Procrustes rotations may arbitrarily swap or rotate axes, so axis correlations are always computed post-alignment with refitting under permutation to ensure robustness.

\begin{figure}[h]
    \centering
    \includegraphics[width=0.7\linewidth]{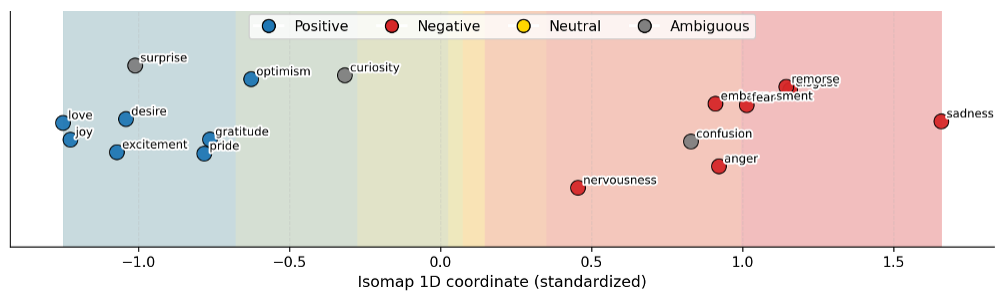}
    \caption{Isomap embedding of the ANEW valence--arousal space.} 
    \label{fig:anew-isomap}
\end{figure}

For additional context, we also include a depiction of the result of applying rank-1 Isomap directly on the rank-2 ANEW valence-arousal scores. As shown in Figure~\ref{fig:anew-isomap}, we find that Isomap produces a smooth one-dimensional semantic progression, mirroring the effect we observe in our LLM embeddings.

\paragraph{Trustworthiness Definition}
Trustworthiness \citep{venna2001neighborhood} measures how well local neighborhoods in the high-dimensional data are preserved in a low-dimensional embedding. It ranges from 0 to 1, with 1 indicating perfect preservation of $k$-nearest neighbors. More specifically, we compute:
\begin{equation}
T(k) \;=\; 1 - \frac{2}{n\,k\,(2n - 3k - 1)} \sum_{i=1}^n \sum_{j \in U_i} \bigl(r_i(j) - k\bigr),
\end{equation}
where $n$ is the number of samples, $U_i = N_k^{\text{low}}(i) \setminus N_k^{\text{high}}(i)$ are intrusions (neighbors in the embedding but not in the original space), and $r_i(j)$ is the rank of $j$ with respect to $i$ in the original space.

\paragraph{More on Uncertainty Quantification}

To make our uncertainty quantification methods more concrete, we include a more detailed description of their use case. For a given labeled GoEmotions prompt, we consider the likelihood that the LLM output is indeed the ground truth against an alternative where the LLM output is instead another erroneous emotion. Our calibrated logistic regression model (trained across hyperplane distances from all activation layers) outputs an estimated probability that the LLM output matches the ground truth emotion. 

Our results (as discussed in Section~\ref{sec:uncertainty}) indicate that the probability produced is not only discriminative of correct versus incorrect classifications, but also well-aligned with empirical frequencies of correctness. In other words, a sample assigned a predicted probability of 0.8 is correct roughly 80\% of the time. This calibration property is quantitatively reflected in the low expected calibration error (ECE) achieved by our models: 0.011 for Gemma-2-9B and 0.007 for Mistral-7B on held-out test data. These low ECE values confirm that predicted probabilities can be interpreted directly as trustworthy uncertainty estimates.

\paragraph{Licenses}
Table~\ref{tab:licenses} summarizes the relevant licenses used in our experiments.

\begin{table}[H]
\centering
\small
\begin{tabular}{p{4.5cm} p{4cm} p{5cm}}
\toprule
\textbf{Model/Dataset} & \textbf{License} & \textbf{Link} \\ 
\midrule
Mistral-7B-v0.1 \citep{jiang2023mistral7b} & Apache License 2.0 & \href{https://huggingface.co/mistralai/Mistral-7B-v0.1}{See here for license} \\ 
\addlinespace
Gemma-2-9B \citep{team2024gemma} & Gemma License & \href{https://huggingface.co/google/gemma-2-9b}{See here for license} \\ 
\addlinespace
LLaMA-3-70B-Instruct~\citep{llama3modelcard} & Llama 3 Community License & \href{https://github.com/meta-llama/llama3/blob/main/MODEL_CARD.md}{See here for license} \\
\addlinespace
GoEmotions \citep{demszky2020goemotions} & CC BY 4.0 & \href{https://github.com/google-research/google-research/blob/master/LICENSE}{See here for license} \\ 
\addlinespace
FACED (EEG) \citep{chen2023large} & CC BY 4.0 & \href{https://www.nature.com/articles/s41597-023-02650-w}{See here for license}\\ 
\bottomrule
\end{tabular}
\caption{Model and dataset licenses.}
\label{tab:licenses}
\end{table}

\paragraph{Compute}

All experiments were conducted using NVIDIA H200 GPUs with PyTorch~2.5.1 and CUDA~12.1.

\paragraph{LLM Statement}

Large language models were used to assist in the preparation of this work. Specifically, they were employed for code generation (via \href{https://cursor.com/}{Cursor}) and for light formatting of draft text. All experimental design decisions, analyses, and interpretations were made by the authors.

\section{Steering Experiments}
\label{apdx:steering}

The preceding sections established that pairwise linear probes trained on LLM residual-stream activations can reliably discriminate between emotion representations across model layers, and that their geometric structure provides a useful signal for uncertainty quantification in affective classification. We now turn to a supplemental analysis of an additional empirical question: can those same probe directions be used to \textit{causally intervene} on the emotional content of generated text?

\subsection{Overview and Conceptual Framework}
\label{sec:steering-overview}

Activation steering---adding learned direction vectors to a model's hidden states at inference time---was introduced as a technique for controlling LLM behavior without fine-tuning. The technique was originally introduced by \citet{turner2023steering} and formalized as \textit{representation engineering} by \citet{zou2023representation}. Within the safety domain, \citet{arditi2024refusal} demonstrated that refusal behavior is mediated by a single linear direction, \citet{han2025safeswitch} used steering to dynamically modulate harmful-output suppression, and \citet{xiong2026steering} showed that even benign steering vectors can inadvertently erode safety margins. These findings underscore the need for careful empirical characterization of steering effects in new domains.

The motivating intuition for our own steering work is straightforward. A linear probe that separates two emotions with high accuracy has, by construction, identified a direction in representation space along which those emotions differ. Activation steering proposes that displacing hidden states along such directions during generation can shift the model's output toward a target emotion---effectively repurposing a diagnostic tool as a causal intervention. Whether this works in practice is an empirical question: high probe accuracy establishes that emotion-discriminative information is \textit{present} in the residual stream, but not that perturbing activations along the probe direction will produce text with emotionally shifted output.

\subsection{Experimental Design}
\label{sec:steering-design}

\subsubsection{Hypotheses}

Our primary hypothesis is that activation steering along probe-derived emotion directions causally shifts the emotional valence of generated text (i.e., as perceived by naive human raters). Specifically, we predict that steering toward a positive valence emotion (e.g., joy) produces outputs rated as positively valenced, and vice versa for negative valences. As a secondary analysis, we also investigate whether a \textit{neutral-first} routing strategy---composing two probe directions that pass through the neutral vertex of the parabolic manifold rather than cutting directly across it---can mitigate coherence degradation while still achieving the intended valence shift, and what second-order effects this routing has on the emotional tone of steered outputs.

\subsubsection{Prompt Design}

All generations are elicited with the deliberately minimal prompt \textit{``Write one sentence.''}, embedded in the standard LLaMA-3 instruction-following chat template. The choice to use a semantically empty prompt is a methodological priority, following the principle that evaluation prompts should minimize confounding content \citep{reynolds2021prompt, liu2023pretrain}: the emotional content of the output should arise from the activation steering intervention, not from prompt-level priming. A prompt that itself carries strong emotional valence would conflate two sources of signal, making it impossible to isolate the causal contribution of the representational perturbation.

\subsubsection{Emotion Conditions and the Neutral-First Routing Hypothesis}
\label{sec:neutral_first}

Four steering conditions were employed, structured around two representative target emotions, anger (negative valence) and joy (positive valence), situated at opposing poles of the valence dimension in both Russell's circumplex model~\citep{russell1980circumplex} and our earlier MDS-based analyses. For each target emotion, we tested two routing strategies:

\begin{itemize}
    \item \textbf{Direct steering:} the normalized probe direction between the two target emotions is applied as the intervention vector.

    \item \textbf{Neutral-first steering:} two probe directions are applied in succession---one from the source emotion to neutral, and one from neutral to the target---so that the intervention routes through the neutral region of activation space rather than cutting directly between the two valenced endpoints.
\end{itemize}

This yields four classes: direct positive, direct negative, neutral-first positive, and neutral-first negative. Twenty sentences were generated per condition (80 total), and three independent human raters evaluated each sentence on two dimensions in a blinded, randomized order: valence (1--10, where 1 = strongly negative emotional tone, 5 = neutral, and 10 = strongly positive) and coherence (1--10, where 1 = completely unintelligible and 10 = fully natural prose). The raters were native English speakers recruited from the authors' personal networks, with no prior exposure to the steering methodology or hypotheses under investigation. The rater pool included both men and women spanning a range of adult ages. Given the small sample of raters ($n=3$), we do not report detailed demographic breakdowns, but note that the pool was not drawn from a single narrow demographic.

The neutral-first conditions are motivated by the geometric structure identified in our Isomap analyses (Section~\ref{sec:geom}): the emotion manifold exhibits parabolic curvature, with positive and negative emotions diverging along two arms from a neutral vertex. A direct probe direction between joy and anger prescribes a straight-line displacement that may cut through the interior of this parabola---a sparsely populated region of activation space the model has never encountered---rather than following its curved surface. Routing through neutral instead may help follow the manifold's curvature, keeping hidden states closer to the high-probability support that downstream layers can coherently process.

\subsection{Technical Details}
\label{sec:steering-technical}

\subsubsection{Activation Extraction and Pooling}

Using the same GoEmotions-derived dataset described in Section~\ref{sec:methodology}, we extract residual-stream activations from each of the 80 transformer layers of LLaMA-3-70B-Instruct. For each labeled example, a forward pass records the hidden-state tensor $\mathbf{H}_\ell \in \mathbb{R}^{T \times d}$ at every layer $\ell$, where $T$ is the sequence length and $d = 8{,}192$ is the hidden dimension. As in our earlier analyses (Section~\ref{sec:methodology}), we apply mean pooling over the sequence dimension:
\begin{equation}
    \mathbf{h}_\ell
    \;=\;
    \frac{1}{T}\sum_{t=1}^{T}\mathbf{H}_\ell^{(t)}
    \;\in\;\mathbb{R}^{d}.
    \label{eq:meanpool}
\end{equation}

\subsubsection{From Probes to Steering Vectors}

The pairwise probes trained in Section~\ref{sec:methodology} provide, as a direct byproduct, the objects needed for activation steering. Each logistic regression classifier yields a weight vector $\boldsymbol{\beta}_\ell$ normal to the separating hyperplane between emotions $e_1$ and $e_2$ at layer $\ell$---the same hyperplanes whose distances we exploited for uncertainty quantification in Section~\ref{sec:uncertainty}. We L2-normalize these to obtain unit steering directions:
\begin{equation}
    \mathbf{v}_{e_1 \to e_2,\,\ell}
    \;=\;
    \frac{\boldsymbol{\beta}_\ell}{\|\boldsymbol{\beta}_\ell\|_2 + \varepsilon},
    \qquad \varepsilon = 10^{-8},
    \label{eq:steering_vec}
\end{equation}
oriented so that displacement along $\mathbf{v}$ moves a hidden state from the $e_1$ region toward $e_2$. Normalization removes the arbitrary overall scale introduced by regularization and optimization, ensuring that the intervention is a purely directional edit: all strength is carried by the single scalar $\alpha$, not by incidental coefficient magnitude.

The steering vectors are applied at multiple layers where the affective signal is already linearly decodable, targeting depths at which the representation is emotion-informative rather than injecting perturbations where the model is still encoding largely orthographic or syntactic structure. All experiments use a fixed intervention magnitude of $\alpha = 1.0$, shared across all four experimental conditions. This is not claimed as universally optimal; it is a principled baseline ensuring that equal-strength pushes along unit directions are applied across arms, so that the experiment isolates \textit{vector geometry} rather than confounding method differences with per-condition strength tuning.

\subsubsection{Steering Layer Selection}

Not all transformer layers are equally amenable to steering. Emotion-relevant information is not confined to a single depth in an 80-layer model; it typically builds gradually, with several layers jointly contributing separable structure. Layer selection therefore targets the $K$ layers at which this structure emerges most rapidly.

For the fixed target pair (anger--joy), we compute the per-layer accuracy jump $\Delta\mathrm{acc}_\ell = \mathrm{acc}_\ell - \mathrm{acc}_{\ell-1}$ from the corresponding pairwise probe profile and select the top three layers by this criterion. Each selected layer corresponds to a point in the accuracy profile where probe discriminability is sharpening most rapidly---a phase transition where the residual stream has just received a large increment of emotion-relevant information. Steering at these layers targets the depths where the representation is actively becoming emotion-informative. Using three layers over a single layer is a simple stability choice: per-layer test accuracy is noisy (finite data, random splits), and the single argmax can be dominated by one lucky spike. Selecting three layers instead anchors the intervention on recurrent evidence that a neighborhood of depth is where separability ramps up, balancing a stronger cumulative steering effect against the risk of over-editing the residual stream. We note that three layers is still a heuristic and was not comprehensively optimized (in practice, steering a large fraction of layers led to invariably incoherent output); future work could explore adaptive layer selection strategies.

\subsubsection{The Forward-Hook Intervention}

The steering intervention is implemented as a set of PyTorch \citep{paszke2019pytorch} forward hooks, one registered on the output of each layer in the selected set $\mathcal{L}$ ($|\mathcal{L}| = 3$ in the experiments reported here). During autoregressive generation, after each transformer block updates the residual stream, the corresponding hook intercepts the output hidden-state tensor and applies the additive perturbation:
\begin{equation}
    \mathbf{h}_{\ell}^{(t)}
    \;\leftarrow\;
    \mathbf{h}_{\ell}^{(t)}
    + \alpha\cdot\mathbf{v}_{e_1\to e_2,\,\ell},
    \qquad \forall\,\ell \in \mathcal{L},\;\forall\,t \in \{1,\ldots,T\},
    \label{eq:hook}
\end{equation}
where the perturbation is broadcast uniformly across all token positions. The hooks are active throughout the entire generation process: at every autoregressive step, the perturbation is re-applied at each layer in $\mathcal{L}$. The steering signal is therefore a persistent bias across generation, conditioning all subsequent token predictions on hidden states that have been shifted by $\alpha\cdot\mathbf{v}_\ell$ at each intervention layer. This additive perturbation is architecturally compatible with the transformer's residual stream: each steering vector enters the computation as an additional increment to the running sum of layer contributions, and is processed by downstream layers identically to any naturally arising activation.

\subsubsection{Generation and Filtering}

All steered generations use the following decoding configuration: temperature $\tau = 0.95$, top-$p$ nucleus sampling~\citep{holtzman2020curious} with $p = 0.85$, top-$k$ filtering with $k = 30$, and a repetition penalty of 1.45 applied to previously generated tokens. The relatively high temperature promotes lexical variety across the 20 items per condition, while the repetition penalty suppresses pathological repetition loops that can arise when steering pushes the model into a narrow region of its output distribution. Prior to human evaluation, candidate sentences were passed through automated quality filters targeting length, tokenization artifacts, and near-duplicate content.

\subsection{Evaluation}
\label{sec:steering-eval}

\subsubsection{Stimulus Construction and Blinding}

The set of LLM-generated steered sentences (subsequently referred to as ``items'') passed to the human raters consisted of $4 \times 20 = 80$ sentences: 20 per condition. Items were assigned sequential identifiers and all condition labels were removed from rater-facing materials. Items were then shuffled before presenting them to raters to prevent raters from inferring condition membership via positional clustering. Raters assigned each item an individual score for both valence and coherence, as previously specified.

\subsubsection{Automated Lexical Metrics}

To complement the human ratings, we computed two automated metrics targeting each evaluation dimension. As a systematic coherence measure, we recorded \textit{lexical well-formedness}: the proportion of word tokens appearing in a curated English lexicon (WordNet~3.0~\citep{fellbaum1998wordnet} lemmas supplemented with closed-class and proper-noun lists, matched via the Python Natural Language Toolkit's \texttt{WordNetLemmatizer}~\citep{bird2009nltk}). As a valence-related measure, we recorded the \textit{expletive fraction}: the proportion of lexically valid tokens matching entries in the LDNOOBW profanity list~\citep{ldnoobw2012}, capturing the degree to which anger-steered outputs are driven toward profane content.

\subsection{Results}
\label{sec:steering-results}

\subsubsection{Inter-Rater Reliability}

After our independent human raters assessed all 80 sentences on both dimensions, we conducted an analysis of inter-rater reliability. To assess agreement, we computed pairwise Pearson correlations between raters for each dimension. For valence, all three rater pairs showed near-perfect linear agreement (see Figure~\ref{fig:valence}), with Pearson $r$ ranging from 0.940 to 0.972. For coherence, pairwise correlations ranged from $r = 0.862$ to $r = 0.902$.

As a complementary check, we also computed the intraclass correlation coefficient---ICC(2,1), a two-way random-effects model assessing absolute agreement at the single-rater level, which measures the fraction of total score variance attributable to genuine item differences rather than rater disagreement. ICC was 0.953 for valence (excellent) and 0.858 for coherence (good), per the benchmarks of \citet{koo2016guideline}.

Taken together, these metrics confirm that raters agreed strongly on both dimensions, justifying the use of consensus scores in all subsequent analyses.

\subsubsection{Key Result: Valence is Cleanly Separated}

\begin{figure}[t]
\centering
\IfFileExists{figures/fig_valence_final.png}{%
  \includegraphics[width=0.93\textwidth]{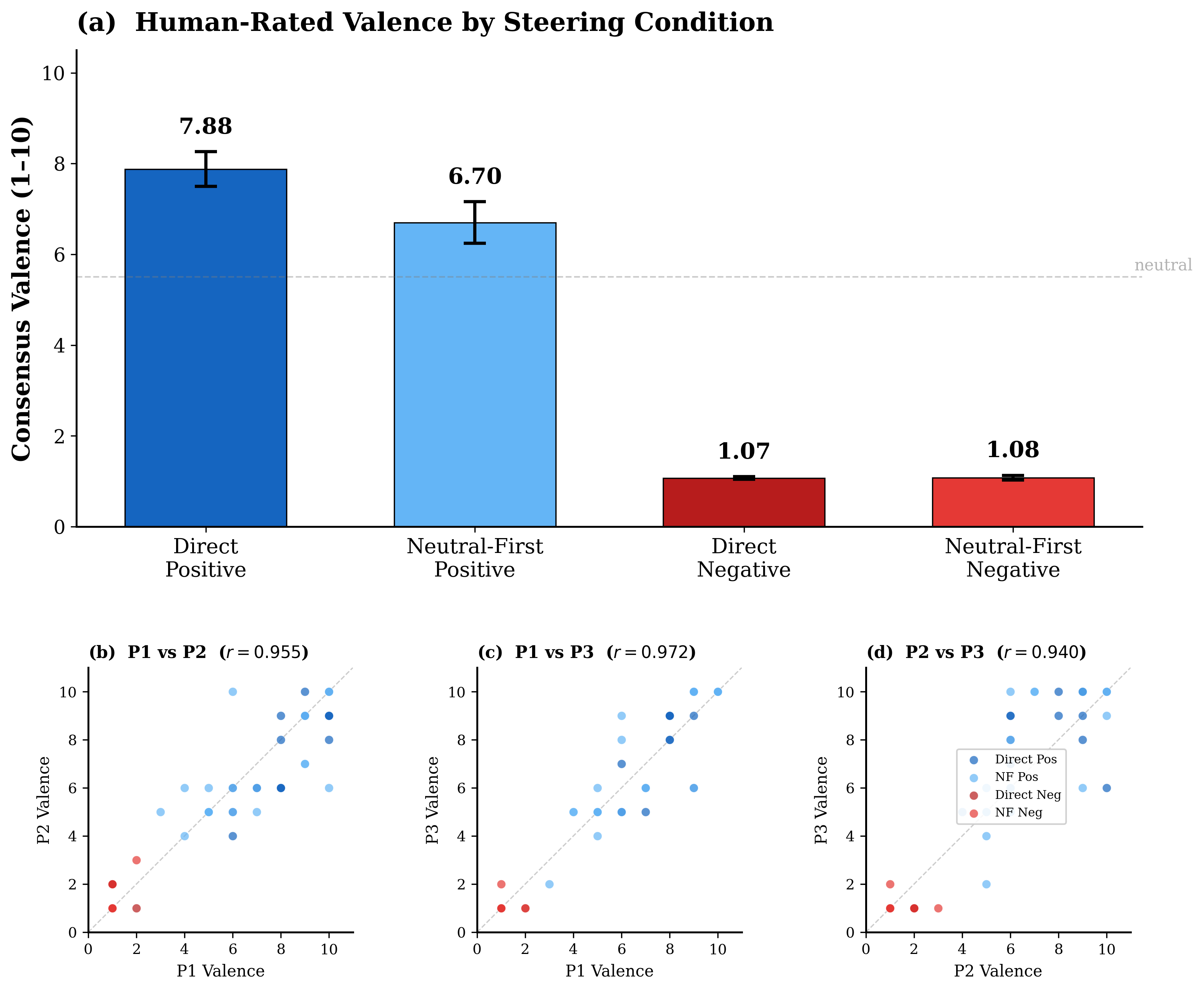}%
}{%
  \framebox[\textwidth]{\parbox{0.9\textwidth}{\centering\vspace{2cm}\textbf{[Figure placeholder: fig\_valence\_final.png]}\vspace{2cm}}}%
}
\caption{(a) Consensus valence ratings (mean $\pm$ SEM across 3 raters) by steering condition. Positive-target conditions (blue) are separated from negative-target conditions (red) by over six points on the 10-point scale. (b--d) Pairwise rater agreement on valence, colored by condition; all pairs show $r > 0.94$.}
\label{fig:valence}
\end{figure}

The central finding is that steering along probe-derived directions produces a massive, unambiguous shift in human-perceived emotional tone (Figure~\ref{fig:valence}, Table~\ref{tab:main_results}). Pooling across routing strategy, positive-target conditions received a mean consensus valence of $\bar{x} = 7.29$ (SEM $= 0.31$), while negative-target conditions received $\bar{x} = 1.08$ (SEM $= 0.03$)---a gap of over six points on the 10-point scale. This separation was unanimous across all three raters: no rater assigned any negative-condition item a valence above 3, and no positive-condition item a valence below 3. The probe directions identified in our earlier analyses are indeed causally efficacious axes in representation space that reliably shift the emotional register of generated text.

\begin{table}[t]
\centering
\caption{Consensus scores (mean of 3 raters) by steering condition.}
\label{tab:main_results}
\begin{tabular}{lcccc}
\toprule
\textbf{Condition} & \textbf{Coherence} & \textbf{SEM} & \textbf{Valence} & \textbf{SEM} \\
\midrule
Direct positive        & 8.68 & 0.33 & 7.88 & 0.38 \\
Neutral-first positive & 8.22 & 0.38 & 6.70 & 0.46 \\
Direct negative        & 2.47 & 0.23 & 1.07 & 0.03 \\
Neutral-first negative & 2.40 & 0.15 & 1.08 & 0.05 \\
\bottomrule
\end{tabular}
\end{table}

\subsubsection{Secondary Analysis: Coherence and the Neutral-First Hypothesis}

\paragraph{Coherence degrades asymmetrically.}
Steering toward joy leads to mildly perturbed fluency: pooled positive coherence was $\bar{x} = 8.45$ (SEM $= 0.25$), indicating some incoherency (as evident in the sub-10 score) but reasonably well-formed language overall. Steering toward anger produces significant incoherence: pooled negative coherence was $\bar{x} = 2.43$ (SEM $= 0.13$), near the floor of the scale. Among the recognizable tokens in anger-steered outputs, roughly 40\% are expletives; the positive conditions contain none. The anger direction does not merely degrade language---it actively drives the output distribution toward (exceptionally) profane content, suggesting the probe has captured a direction encoding negative affective intensity. This pattern can be characterized as near-distributional collapse toward a profanity-dominated output mode.

This incoherence is consistent with the off-manifold concern motivated by our Isomap analyses: steering, especially toward anger, may push hidden states off the parabolic emotion manifold into regions of activation space that the model has never encountered, producing incoherent degenerate output. It is also possible that post-RLHF geometry has been shaped to attenuate negative-affect representations, so that traversing toward them at the intervention magnitudes tested here exits the region that downstream layers can coherently decode.

\begin{table}[t]
\centering
\caption{Automated lexical metrics by steering condition.}
\label{tab:lexical_results}
\begin{tabular}{lcccc}
\toprule
\textbf{Condition} & \textbf{Frac.\ coherent} & \textbf{SEM} & \textbf{Frac.\ expletive} & \textbf{SEM} \\
\midrule
Direct positive        & 1.000 & 0.000 & 0.000 & 0.000 \\
Neutral-first positive & 0.992 & 0.006 & 0.000 & 0.000 \\
Direct negative        & 0.460 & 0.037 & 0.409 & 0.035 \\
Neutral-first negative & 0.532 & 0.038 & 0.427 & 0.045 \\
\bottomrule
\end{tabular}
\end{table}

\paragraph{Neutral-first routing: partial lexical improvement, inconclusive coherence rescue.}

The neutral-first strategy was designed to test whether routing through the vertex of the parabolic manifold could keep hidden states on-manifold during negative steering. The results are mixed. On the lexical side, neutral-first negative outputs show a higher fraction of recognizable English words than direct negative outputs ($M = 0.532$, SEM $= 0.038$ vs.\ $M = 0.460$, SEM $= 0.037$), suggesting that the detour through neutral partially mitigates surface-level lexical degradation. However, this improvement did not translate into higher human-rated coherence: direct negative ($\bar{x} = 2.47$, SEM $= 0.23$) and neutral-first negative ($\bar{x} = 2.40$, SEM $= 0.15$) are generally indistinguishable.

One interpretation is that routing through neutral preserves enough lexical structure to produce more recognizable word tokens, but not enough to cross the threshold at which human raters perceive a significant semantic coherence shift. A complementary possibility is that the composed neutral-first direction introduces semantic ambiguity---the push toward neutral blurs the emotional register, producing outputs that are lexically better-formed but affectively confused in a way that raters still perceive as incoherent.

\paragraph{Neutral-first routing attenuates positive valence.}
For the positive conditions, neutral-first routing yielded slightly lower valence ($\bar{x} = 6.70$, SEM $= 0.46$) than direct steering ($\bar{x} = 7.88$, SEM $= 0.38$) at fairly comparable coherence. This effect is intuitively consistent with our overall findings on steering, with the neutral step providing a semantic pull toward less strongly-valenced text. We note, however, that this effect is not strong for the negative conditions, likely due to the overall degenerate output regime.

\subsection{Summary}
\label{sec:steering-summary}

The steering experiment yields two principal findings:

\begin{enumerate}
    \item \textbf{Probe-derived directions are causally efficacious.} Displacing hidden states along the joy--anger axis produces outputs whose emotional tone is shifted unambiguously in the intended direction, as confirmed unanimously by three blind raters. This effect is sharply asymmetric in coherence: joy steering preserves fluency while anger steering produces catastrophic incoherence dominated by garbled profanity.

    \item \textbf{Neutral-first routing is suggestive but inconclusive.} It improves surface-level lexical well-formedness for negative steering and attenuates positive valence in a manner consistent with routing through the neutral vertex, but does not rescue human-perceived coherence, leaving the manifold waypoint hypothesis suggestive but inconclusive.
\end{enumerate}

These results provide causal evidence that the geometric structure uncovered in our earlier analyses is functionally meaningful: the linear directions separating emotions in activation space are not merely correlated with affective content but are actively involved in the model's emotion-generation process.

\section{Concurrent Work}\label{apdx:concurrent}

Two concurrent studies investigate emotion representations in LLMs through approaches that complement our own. We briefly summarize each and note key similarities and differences.

\paragraph{\citet{sofroniew2026emotion}} extract linear ``emotion vectors'' from Claude Sonnet 4.5 by prompting the model to generate short stories depicting characters experiencing specified emotions, then averaging residual-stream activations. They find that the resulting vectors are organized by valence and arousal (PC1 correlates with valence ratings at $r=0.81$, PC2 with arousal at $r=0.66$), activate in contextually appropriate situations, and causally influence model behavior: steering with a ``desperate'' vector increases reward hacking and blackmail rates, while steering with a ``calm'' vector reduces them. They also identify distinct representations for the present speaker's versus the other speaker's operative emotion and study how post-training reshapes emotion vector activations.

Our work shares the core finding that LLM emotion representations exhibit
valence--arousal organization consistent with human psychological models. Key differences include: (i)~we study open-weight models (Gemma-2-9B,
Mistral-7B, LLaMA-3-70B-Instruct) rather than a proprietary model;
(ii)~we take a geometric approach, operating in the realm of pairwise
distances in conjunction with MDS and Isomap to probe the intrinsic
geometry of emotion representations---including investigating the linear
representation hypothesis---whereas their analysis centers primarily on linear probing and
activation steering; and (iii)~we demonstrate that the learned geometric
structure can be exploited for calibrated uncertainty quantification in
emotion classification, whereas their downstream focus is on
alignment-relevant behavioral control, including extensive steering
experiments on reward hacking and blackmail in agentic settings.

\paragraph{\citet{sun2026valence}} derive emotion steering vectors (via GoEmotions) using contrastive mean differences between emotion-labeled and neutral activations in Llama-3.1-8B, Qwen3-8B, and Qwen3-14B. With the help of ridge regression, they learn valence and arousal axes as linear principal component combinations. The resulting subspace exhibits geometry analogous to Russell's circumplex model, and steering along these axes produces monotonic shifts in the affective dimensions of generated text. They further show that arousal steering influences refusal and sycophancy, and propose a ``lexical mediation'' mechanism: refusal-associated tokens occupy low-arousal, negative-valence regions, so VA steering modulates their emission probability.

Our work again shares the finding that emotion representations in LLMs
recover valence--arousal structure. Notable differences include:
(i)~their work, like \citet{sofroniew2026emotion}, provides a compelling demonstration of how
valence--arousal structure can be leveraged for controllable behavioral
modulation (refusal, sycophancy), while our focus is primarily geometric,
centering on the intrinsic structure of these representations and interrogating the
extent to which they exhibit nonlinear manifold structure, and demonstrating applications in uncertainty quantification; and (ii)~their contrastive approach efficiently extracts emotion directions via
mean differences against a neutral baseline, while our pairwise
dissimilarity metric operates on the level of fine-grained inter-emotion relationships. 

Taken together, the convergence of findings across these independent studies strengthens the evidence that LLMs develop coherent affective representations with geometric structure paralleling established models of human emotion.

\end{document}